%% file: main.tex
\documentclass{article}
\input{math_commands}

\usepackage[final]{corl_2019} 
\usepackage{symbols}
\usepackage{epsfig}
\usepackage[normalem]{ulem}
\usepackage{amsmath}
\usepackage{amssymb}
\usepackage{symbols}
\usepackage{algorithm}
\usepackage{algorithmic}
\usepackage{ctable} 
\usepackage{pifont}
\usepackage{bbm}
\usepackage{framed}
\usepackage{lipsum} 
\usepackage{float}
\usepackage{multirow}
\usepackage[font=small]{caption}
\usepackage{subcaption}
\usepackage{animate}
\usepackage{wrapfig}
\usepackage{multicol}
\usepackage{makecell}
\usepackage{enumitem}
\usepackage{adjustbox}

\usepackage[symbol]{footmisc}

\title{PIC: Permutation Invariant Critic for Multi-Agent Deep Reinforcement Learning}

\author{
  Iou-Jen Liu\thanks{Authors contributed equally.} \qquad Raymond A. Yeh\footnotemark[1]\qquad Alexander G. Schwing\\
  University of Illinois at Urbana-Champaign\\
  United States\\
  \texttt{\{iliu3, yeh17, aschwing\}@illinois.edu} \\
}

\begin{document}
\maketitle


\begin{abstract}
\input{abs}
\end{abstract}


\input{intro}
\input{rel}
\input{back}
\input{app}

\input{exp}
\input{conc}

\input{ack}


{
\setlength{\bibsep}{1.4pt}
\bibliography{example}
}

\clearpage
\input{supp}

\end{document}

%% file: math_commands.tex
\usepackage{amsmath,amsfonts,bm}

\newcommand{\cO}{{\cal O}}

\newcommand{\expectation}{{\mathbb{E}}}

\def\Eqref#1{Equation~\ref{#1}}

\def\1{\bm{1}}

\def\vs{{\bm{s}}}

\DeclareMathAlphabet{\mathsfit}{\encodingdefault}{\sfdefault}{m}{sl}
\SetMathAlphabet{\mathsfit}{bold}{\encodingdefault}{\sfdefault}{bx}{n}



%% file: abs.tex
Sample efficiency and scalability to a large number of agents are two important goals for multi-agent reinforcement learning systems. Recent  works got us closer to those goals, addressing non-stationarity of the environment from a single agent's perspective  by utilizing a  deep net critic which depends on all observations and actions.
The critic input concatenates agent observations and actions in a user-specified order. However, since deep nets aren't permutation invariant, a permuted input changes the critic output despite the environment remaining identical.
To avoid this inefficiency, we propose a 
`permutation invariant critic' (PIC), which yields identical output irrespective of the agent permutation.
This consistent representation enables our model to scale to 30 times more agents and to achieve improvements of test episode reward between 15\% to 50\% on the challenging multi-agent particle environment (MPE). 

%% file: intro.tex
\section{Introduction}
\label{sec:intro}
Single-agent deep reinforcement learning
has achieved impressive performance in many domains, including playing Go~\cite{Silver16, Silver17} and Atari games~\cite{dqn1, dqn2}.  However, many real world problems, such as traffic congestion reduction~\cite{Bazzan08, Sunehag18}, antenna tilt control~\cite{Dandanov17}, and dynamic resource allocation~\cite{Nguyen18} are more naturally modeled as multi-agent systems. Unfortunately, directly deploying single-agent reinforcement learning to each agent in a multi-agent system does not result in satisfying performance~\cite{Tang93, maddpg}.
Particularly, in multi-agent reinforcement learning~\cite{Nguyen18, maddpg, Foerster17, Foerster18, Iqbal19, Jiang18, Das19,  Foerster16, Kim19, Shu19, Han19}, estimating the value function is challenging, because the environment is non-stationary from the perspective of an individual agent~\cite{maddpg, Foerster17}. To alleviate the issue, recently,
multi-agent deep deterministic policy gradient (MADDPG)~\cite{maddpg} proposed a centralized critic whose input is the concatenation of all agents' observations and actions. Similar to MADDPG,~\citet{Foerster17, Foerster18, Kim19, Jiang18, Das19, Iqbal19, Yang18} also deploy centralized critics to handle a non-stationary environment.

However, concatenating all agents' observations and actions assigns an implicit order, 
\ie, the placement of an agent's observations and actions will make a difference in the predicted outcome.
Consider the case of two {\em homogeneous} agents and let us denote the action and observation of the two agents with `A' and `B.' There exists two equally valid permutations (AB) and (BA) which represent the environment. Using a permuted input in classical deep nets will result in a different output.
Consequently, referring to the same underlying state of the environment with two different  vector representations makes  learning of the critic sample-inefficient: the deep net needs to learn that both representations are identical.
Due to an increase in the number of possible permutations, this representational inconsistency worsens as the number of agents grows.

To address this concern, we propose the `permutation invariant critic' (PIC). Due to the permutation invariance property of PICs,  the same environment state will result in the same critic output, irrespective of the agent ordering (as shown in \figref{fig:idea_fig}). 
In addition, to tackle environments with  homogeneous and heterogeneous agents (\eg, agents which have different action space, observation space, or play different roles in a task), we augment PICs  with  attributes. This enables the proposed PIC to model the relation between heterogeneous agents. 

For rigorous results we follow the strict evaluation protocol proposed by~\citet{Henderson17} and~\citet{Colas18} when performing experiments in multi-agent particle environments (MPEs)~\cite{maddpg, Mordatch17}.  We found that  permutation invariant critics result in $15\%$ to $50\%$ higher average test episode rewards than the MADDPG baseline~\cite{maddpg}. Furthermore, we scaled the MPE to 200 agents. Our permutation invariant critic successfully learns the desired policy in environments with a large number of agents, while the baseline MADDPG~\cite{maddpg} fails to develop any useful strategies. 

In summary, our main contributions are as follows: a) We develop a permutation invariant critic (PIC) for multi-agent reinforcement learning algorithms. Compared with classic MLP critics, the PIC achieves  better sample efficiency and scalability. b) To deal with heterogeneous agents we study adding  attributes. c) We speedup the multi-agent particle environment (MPE)~\cite{maddpg, Mordatch17} by a factor of 30. This permits  to scale the number of agents to 200, 30 times more than those used in the original MPE environment (6 agents). {Code is available at \url{https://github.com/IouJenLiu/PIC}.}

\input{intro_fig}

%% file: intro_fig.tex
\begin{figure}
\vspace{-1cm}
\centering
\includegraphics[height=5.5cm]{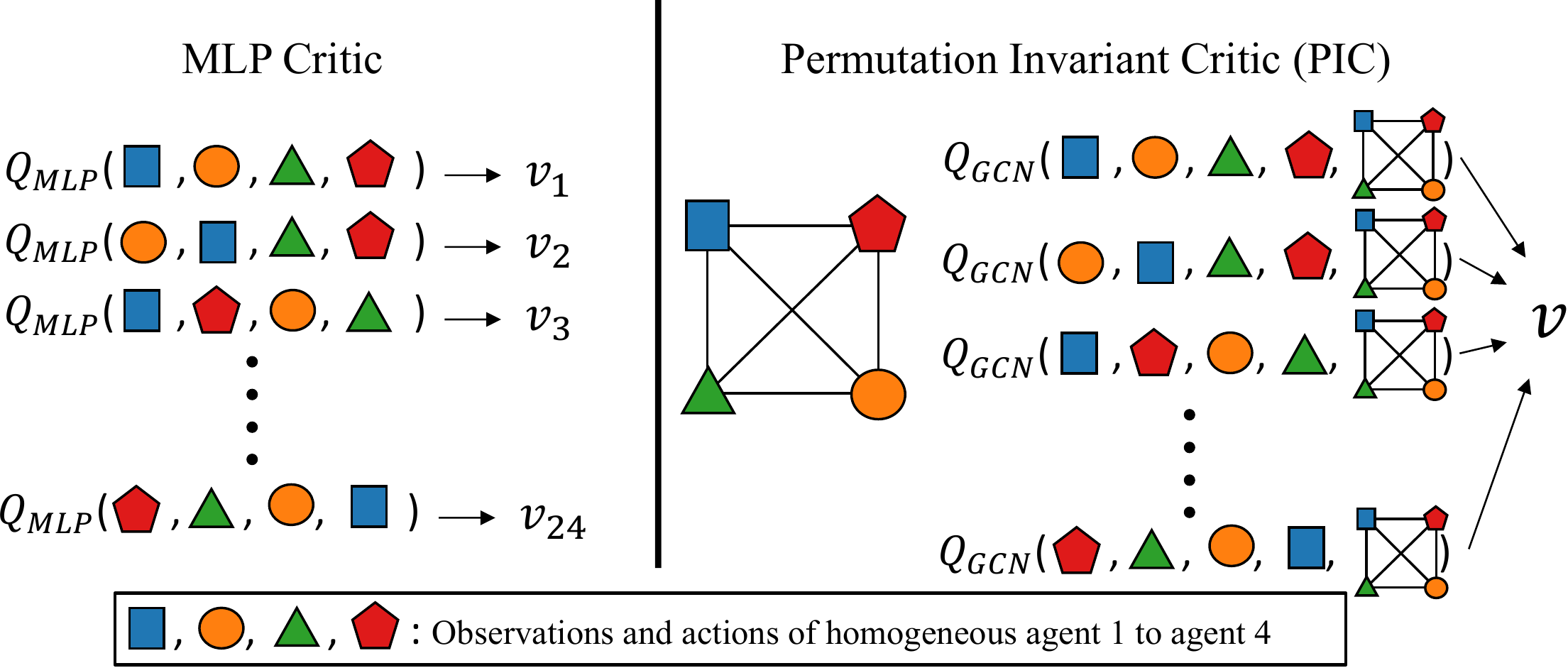}
\vspace{-0.2cm}
\caption{Consider an environment with four homogeneous cooperative agents. Different permutations of the agents' observations and actions refer to the same underlying environment state. However, MLP critics result in $4!$ different outputs for the same environment state. In contrast, permutation invariant critics yield the same output value for all $4!$ equivalent permutations.
}
\label{fig:idea_fig}
\vspace{-0.6cm}
\end{figure}

%% file: rel.tex

\section{Related Work}
We briefly review 
graph neural nets, which are the building block of PICs, and multi-agent deep reinforcement learning algorithms with centralized critics. 

\noindent{\bf Graph Neural Networks.}
Graph neural networks are deep nets which operate on graph structured data~\cite{scarselli2009graph}. Input to the network are hence a set of node vectors and connectivity information about the nodes. More notably, these graph networks are permutation equivariant, \ie, the ordering of the nodes in a vector representation does not change the underlying graph~\cite{zaheer2017deep}. Many variants of graph networks exists, for example, Graph Convolutional Nets (GCN)~\cite{kipf2017semi}, the Message Passing Network~\cite{gilmer2017neural}, and  others~\cite{zaheer2017deep, qi2017pointnet}. The relation and difference between these approaches are reviewed in~\cite{battaglia2018relational}. The effectiveness of graph nets has been shown  on tasks such as link prediction~\cite{schlichtkrull2018modeling, zhang2018link}, node classification~\cite{hamilton2017inductive, kipf2017semi}, language and vision~\cite{NarasimhanNIPS2018,SchwartzCVPR2019a}, graph classification~\cite{yeh2019diverse,duvenaud2015convolutional, zhang2018end, ying2018hierarchical}, \etc.
Graph nets also excel on point sets, \eg,~\cite{zaheer2017deep, qi2017pointnet}. 
Most relevant to our multi-agent reinforcement learning setting, graph networks have been shown to be effective in modeling and reasoning about physical systems~\cite{battaglia2016interaction} and multi-agent sports dynamics~\cite{hoshen2017vain, kipf2018neural, yeh2019diverse}. 
Different from these works, here, we study the effectiveness of graph nets for multi-agent reinforcement learning.

\noindent{\bf Multi-agent Reinforcement Learning.} 
To deal with non-stationary environments form the perspective of a single agent, MADDPG~\cite{maddpg} uses a centralized critic that operates on all agents' observations and actions. 
Similar to MADDPG,~\citet{Foerster18} use a centralized critic. In addition, to handle the credit assignment problem~\cite{Nguyen18, Panait05, Chang03}, a counterfactual baseline has been proposed to marginalize one agent's action and keep the other agents' actions fixed. 
In ``Monotonic Value Function Factorisation'' (QMIX)~\cite{qmix} each agent maintains its own value function which conditions only on the agent's local observation. The overall value function is estimated via a non-linear combination of an individual agent's value function. \citet{Iqbal19} propose an attention mechanism which enables the centralized critic to select relevant information for each agent. However, as discussed in~\secref{sec:intro}, the output of centralized critics parameterized by classic deep nets differs if the same environment state is encoded with a permuted vector. This makes learning inefficient. 

``Graph convolutional RL'' (DGN)~\cite{Jiang19}
is concurrent work on arXiv which uses a nearest-neighbor graph net as the Q-function of a
deep Q-network (DQN)~\cite{dqn1, dqn2}. However, the nearest-neighbor graph net only has access to local information of an environment. Consequently, the Q-function in DGN is not a fully centralized critic. Therefore, it  suffers from the non-stationary environment issue~\cite{maddpg, Foerster17}. In addition, due to the DQN formulation, DGN can only be used in environments with discrete action spaces. Note, DGN considers  homogeneous cooperative agents and leaves environments with heterogeneous cooperative agents to future work. In contrast, our permutation invariant critic is fully centralized and can be scaled to a large number of agents. Thanks to different node attributes, our approach can handle environments with heterogeneous cooperative agents. In addition, our approach is designed for continuous state and action spaces.

%% file: back.tex

\section{Preliminaries}
\label{sec:back}

\subsection{\bf Deep Deterministic Policy Gradient}
In classic single-agent reinforcement learning, an agent interacts with the environment and collects rewards over time. Formally, at each timestep $t \in \{1,\ldots, H \}$, with $H$  the horizon, the agent finds itself in state $s^t \in \cS$ and selects an action $a^t \in \cA$ according to a deterministic policy $\mu$.  Hereby, $\cS$ is the state space, and $\cA$ is the action space. Upon executing an action, the agent  runs into the next state $s^{t+1} \in \cS$ and obtains a scalar reward $r^t \in  \mathbb{R}$. Given a trajectory $\{s^t, a^t, r^t\}_{t=1}^H$ of length $H$ collected by following a policy, we obtain the discounted return $R = \sum_{t=1}^H\gamma^tr^t$, where $\gamma \in (0, 1]$ is the discount factor. The goal of reinforcement learning is to find a policy which maximizes the return $R$. 

Deep deterministic policy gradient (DDPG)~\cite{ddpg} is a widely used deep reinforcement learning algorithm for continuous control. In DDPG, 
a deterministic policy $\mu(s^t|\theta^\mu)$, which is parameterized by $\theta^\mu$, maps a state $s^t$ to an action $a^t$. A critic $Q(s^t, a^t|\theta^Q)$, which is parameterized by $\theta^Q$, is deployed to estimate the return of taking action $a^t$ at state $s^t$. The parameters $\theta^\mu$ of the actor policy are updated iteratively so that the action $a^t=\mu(s^t|\theta^\mu)$ maximizes the critic $Q(s^t, a^t|\theta^Q)$, \ie,
\begin{equation}
\max_{\theta^\mu} J(\theta^\mu) \quad\text{with}\quad J(\theta^\mu)  :=  \expectation_{s^t\sim \cD}[Q(s^t, a^t|\theta^Q)|_{a^t=\mu(s^t|\theta^\mu)}].
\end{equation}
Here $s^t$ is drawn from the replay buffer $\cD$ which stores experience tuples $(s^t, a^t, r^t, s^{t+1})$, \ie, $\cD = \{(s^t, a^t, r^t, s^{t+1})\}_t$. Using the chain rule, the gradient is
\begin{equation}
 \nabla_{\theta^\mu} J(\theta^\mu) = \expectation_{s^t\sim \cD}[\nabla_{\theta^\mu}Q(s^t, a^t|\theta^Q)|_{a^t=\mu(s^t|\theta^\mu)}] \nonumber 
                                                              = \expectation_{s^t\sim \cD}[\nabla_{\theta^\mu} \mu(s^t|\theta^\mu)\nabla_{a^t} Q(s^t, a^t|\theta^Q)|_{a^t=\mu(s^t|\theta^\mu)}]. \nonumber
\end{equation}

To  optimize \wrt $\theta^Q$, similar to deep Q-learning~\cite{dqn1, dqn2}, we minimize the loss 
  \begin{equation}
 L(\theta^Q) = \expectation_{(s^t, a^t, r^t, s^{t+1})\sim \cD}[(Q(s^t, a^t|\theta^Q) - y^t )^2]. 
 \end{equation}
 The experience tuple $(s^t, a^t, r^t, s^{t+1})$  is drawn from the replay buffer $\cD$ and the target value $y^t$ is defined as 
  \begin{equation}
 y^t = r^t + \gamma Q^-(s^{t+1}, a^{t+1}|\theta^{Q^-})|_{a^{t+1}=\mu(s^{t+1})},
 \end{equation}
 where  $Q^-$ is a recent Q-network parameterized by a past $\theta^{Q^-}$.
  
\subsection{\bf Multi-agent Markov Decision Process}
To extend single-agent reinforcement learning to the multi-agent setting, we first define the multi-agent Markov decision process (MDP). We consider partially observable multi-agent MDPs~\cite{Littman94}. An $N$-agent partially observable multi-agent MDP is defined by a transition function ${\cal T}$, a set of reward functions $\{{\cal R}_1 ,\ldots, {\cal R}_N\}$, a state space $\cS$, a set of observation spaces $\{{\cal O}_{1} \ldots, {\cal O}_N\}$, and a set of action spaces $\{\cA_1, \ldots, \cA_N\}$. Action space $\cA_i$, observation space $\cO_i$ and reward function ${\cal R}_i$ correspond to agent $i \in \{1,\ldots, N\}$. 
The transition function ${\cal T}$ maps the current state and the actions taken by all the agents to a next state,  \ie, ${\cal T}: \cS \times \cA_1 \times \ldots \times \cA_N \rightarrow \cS$. Each agent receives reward ${\cal R}_i: \cS \times \cA_1 \ldots \times \cA_N \rightarrow  \mathbb{R}$ and 
observation $o_i$ that is related to the state, \ie, $o_i: \cS \rightarrow {\cal O}_i$. 

The goal of agent $i$ is to maximize the expected return $\sum_{t=0}^H \gamma^tr_i^t$. Note that, the goal of $N$ cooperative agents  is to maximize the collective expected return $\sum_{t=0}^H \gamma^t(\sum_{j =1}^N r_j^t)$.

\subsection{\bf Multi-agent Deep Deterministic Policy Gradient}

In this paper, we study multi-agent reinforcement learning using the decentralized execution and centralized training framework~\cite{Foerster16, maddpg}. Multi-agent deep deterministic policy gradient (MADDPG)~\cite{maddpg} is a well-established algorithm for this framework. 
Consider $N$ agents with policies $\{\mu_1, \ldots, \mu_N\}$, which are parameterized by $\{\theta^{\mu}_1, \ldots, \theta^{\mu}_N \}$. The {\em centralized} critics $\{Q_1, \ldots, Q_N\}$ associated with the $N$ agents are parameterized by $\{\theta_1^Q, \ldots, \theta_N^Q\}$. Following DDPG, the parameters $\theta^{\mu}_i$ for policy $\mu_i$ are updated iteratively so that the associated critic $Q_i$ is optimized via
\begin{equation}
\max_{\theta^\mu_i} J(\theta^\mu_i) \quad\text{with}
~J(\theta_i^\mu)  :=  \expectation_{(\boldsymbol{x}^t, \boldsymbol{a}^t)\sim \cD}[Q_i(\boldsymbol{x}^t,  a_1^t, \ldots, a_N^t|\theta^Q_i)|_{a_i^t=\mu_j(o_i^t|\theta_i^\mu)}],
\label{eq:maddpg_j}
\end{equation}
where $\boldsymbol{x}^t$ and $\boldsymbol{a}^t$ are the concatenation of all agents' observation and action at timestep $t$, \ie, $\boldsymbol{x}^t = (o^t_1, \ldots, o^t_N)$ and $\boldsymbol{a}^t = (a^t_1, \ldots, a^t_N)$.  
Note, $o_i^t$ is the observation received by agent $i$ at time step $t$.  
Using the chain rule, the gradient is derived as follows:
\begin{equation}
\nabla_{\theta_{i}^\mu} J(\theta_{i}^\mu)  = 
                                                            \expectation_{(\boldsymbol{x}^t, \boldsymbol{a}^t)\sim \cD}[\nabla_{\theta_{i}^\mu} \mu_{i}(o_i^t|\theta_i^\mu))\nabla_{a^t_i}Q_i(\boldsymbol{x}^t, a_1^t, \ldots, a_N^t|\theta^Q_i)|_{a_i^t=\mu_{i}(o_i^t|\theta_{i}^\mu)}].
\end{equation} 
Following DDPG, the centralized critic parameters $\theta_i^Q$ are optimized by minimizing the loss
\begin{equation}
 L(\theta^Q_{i}) = \expectation_{(\boldsymbol{x}^t, \boldsymbol{a}^t, \boldsymbol{r}^t, \boldsymbol{x}^{t+1})\sim \cD}[(Q_{i}(\boldsymbol{x}^t, \boldsymbol{a}^t|\theta^Q_{i}) - y_{i}^t )^2], 
  \label{eq:maddpg_q_loss}
 \end{equation}
where $ \boldsymbol{r}^t$ is the concatenation of rewards received by all agents at timestep $t$, \ie, $\boldsymbol{r}^t=(r_1^t, \ldots, r_N^t)$.  The target value $y_{i}^t$ is defined as follows
 \begin{equation}
y_{i}^t = r_i + \gamma Q_{i}^-(\boldsymbol{x}^{t+1}, a_1^{t+1}, \ldots, a_N^{t+1}|\theta^{Q^-}_{i})|_{a_j^{t+1}=\mu_{j}(o^{t+1}_i),~j=1,\ldots,N},
 \label{eq:maddpg_y}
\end{equation}
 where  $Q_i^-$ is a Q-network parameterized by a past $\theta^{Q^-}_i$.

%% file: app.tex

\section{Permutation Invariant Critic (PIC) and Environment Improvements}
We first describe the proposed permutation invariant critic (PIC), then show improvements for MPE. 

\subsection{Permutation Invariant Critic (PIC)}

Consider training $N$ homogeneous cooperative agents using a centralized critic $Q$~\cite{Nguyen18, maddpg, Foerster17, Foerster18, Iqbal19, Jiang18, Das19,  Foerster16, Kim19, Shu19, Han19}. As discussed in~\secref{sec:back}, the input to the centralized critic $Q$ is the concatenation of all agents observations and actions.  Let $\boldsymbol{x}^t \in \mathbb{R}^{N \times K_o}$ denote the concatenation of all agents' observations at timestep $t$, where $K_o$ is the dimension of each observation $o_i^t$.
Similarly, we represent the concatenation of all agents' actions at timestep $t$ via $\boldsymbol{a}^t \in \mathbb{R}^{N \times K_a}$, where $K_a$ is the dimension of each agent's action $ a_i^t$. 
Note that  concatenating observations and actions implicitly imposes an agent ordering. Any agent ordering seems plausible. 
Importantly, shuffling the agents observations and actions doesn't change the state of the environment. 
One would therefore expect the centralized critic $Q$ to return the same output if the input is concatenated in a different order. Formally, this property is called \emph{permutation invariance}, \ie, we strive for a critic such that
$$
Q(M_i \boldsymbol{x}^t, M_i \boldsymbol{a}^t| \theta^Q) = Q(M_j \boldsymbol{x}^t, M_j \boldsymbol{a}^t|  \theta^Q) \;\;\; \forall M_i, M_j \in \cP,
$$
where $M_i$ and $M_j$ are two permutation matrices from the set of all possible permutation matrices $\cP$. 

To achieve permutation invariance, we propose to use a graph convolutional neural net (GCN) as the centralized critic. 
In the remainder of the section, we describe the GCN model in detail and discuss how to deploy the permutation invariant critic to environments with homogeneous and heterogeneous agents. 

\noindent{\bf Permutation Invariant Critic.}
We model an $N$-agent environment as a graph. Each node represents an agent, and the edges capture the relations between agents. 
To compute the scalar output of the critic we use 
$L$ graph convolution layers $\{f^{(1)}_\text{GCN}, \ldots, f^{(L)}_\text{GCN}\}$. A graph convolution layer takes node representations and the graph's adjacency matrix as input and computes a new representation for each node. More specifically,  $f^{(l)}_{\text{GCN}}$ maps the input ${\bold h^{(l-1)}} \in \mathbb{R}^{N \times K^{(l)}_{\text{in}}}$ to the output ${\bold h^{(l)}} \in \mathbb{R}^{N \times K^{(l)}_{\text{out}}}$, where $K^{(l)}_{\text{in}}$ and $K^{(l)}_{\text{out}}$ are the input and output node representation's dimension for layer $l$. Formally,
\begin{equation}
{\bold h}^{(l)} = f^{(l)}_{\text{GCN}}({\bold h}^{(l-1)}) :=  \sigma\left(\frac{1}{N} A_{\text{adj}}{\bold h^{(l-1)}}W^{(l)}_{\text{other}} + {\text{diag}(1)}_{N}{\bold h^{(l-1)}}W^{(l)}_{\text{self}}\right),
\label{eq:GCN}
\end{equation}
where $A_{\text{adj}}$ is the graph's adjacency matrix, $W^{(l)}_{\text{self}}, W^{(l)}_{\text{other}} \in \mathbb{R}^{K^{(l)}_{\text{in}} \times K^{(l)}_{\text{out}}}$
are the layer's trainable weight matrices, $\sigma$ is an element-wise non-linear activation function, and $\text{diag}(1)_{N}$ is an identity matrix of size $N\times N$.

Note that in~\Eqref{eq:GCN}, each agent's representation, \ie, each row of ${\bold h}^{(l)}$, is multiplied with the same set of weights $W_{\text{other}}$ and $W_{\text{self}}$. 
Due to this weight sharing scheme, 
a permutation matrix $M$ applied at the input is equivalent to applying it at the output, \ie, 
\begin{equation}
f^{(l)}_{\text{GCN}}(M {\bold h}) = M f^{(l)}_{\text{GCN}} ({\bold h}).
\end{equation}
Another advantage of the weight sharing scheme is that the number of trainable parameters of PIC does not increase with the number of agents.

Subsequently, 
a pooling layer is applied to the $L$-th graph convolutional layer's representation ${\bold h}^{(L)}$. 
Pooling is performed over the agents' representation, \ie, over the rows of  ${\bold h}^{(L)}$. We refer to the output of the pooling layer as $\bold v \in  \mathbb{R}^{K^{(L)}_\text{out}}$. 
Either average pooling or max pooling is suitable. Average pooling, subsequently denoted $f_{\text{avg}}$,  averages the node representations, \ie, ${\bold v} = \frac{1}{N}\sum_{i=1}^{N} {\bold h}^{(L)}_i$. Max pooling, subsequently referred to as $f_{\text{max}}$, takes the maximum value across the rows for each of the columns,
\ie, ${\bold v}_{j} = \max_{i}{{\bold h}^{(L)}_{i, j}}$ $\forall j \in \{1, \ldots, K^{(L)}_\text{out}\}$.
Both max pooling and average pooling satisfy
the permutation invariance property as summation and element-wise maximization are commutative operations. Therefore, an $L$-layer graph convolutional net is obtained via $f_{\text{max}} \circ f^{(L)}_{\text{GCN}} \circ \ldots \circ  f^{(1)}_{\text{GCN}}$.

\noindent {\bf Homogeneous setting.}
If agents in an environment are homogeneous, 
we first concatenate the observations $\boldsymbol{x}^t$ and actions $\boldsymbol{a}^t$ into a matrix $\boldsymbol{z}^t$, \ie, $\boldsymbol{z}^t := [\boldsymbol{x}^t, \boldsymbol{a}^t] \in \mathbb{R}^{N \times (K_o + K_a)}$. Setting $K_\text{in}^{(1)} = K_o + K_a$ and $h^{(0)}=\boldsymbol{z}^t$, we construct a permutation invariant critic $Q_\text{PIC}$ as follows:
\begin{equation}
 Q_\text{PIC}(\boldsymbol{z}^t) : = f_{\text{v}} \circ f_{\text{max}} \circ f^{(L)}_{\text{GCN}} \circ \ldots \circ  f^{(1)}_{\text{GCN}}(\boldsymbol{z}^t).
 \label{eq:q_homo}
\end{equation}
Hereby $f_{\text{v}}$ maps the output of the graph nets to a real number, which is the estimated scalar critic value for the environment observation $\boldsymbol{x}^t$ and action $\boldsymbol{a}^t$. 
We model $f_{\text{v}}$ with a standard fully connected layer which maintains permutation invariance. 
To ensure that the permutation invariant critic is fully centralized, \ie, to ensure that we consider all agents' actions and observations, we use an adjacency matrix corresponding to   a complete graph, \ie,  $A_{\text{adj}}$ is a matrix of all ones with zeros on the diagonal. 
As later mentioned in \secref{sec:exp} we also study other settings but found a complete graph to yield best results.

\noindent {\bf Heterogeneous setting.}
Consider $N$ cooperative agents, which are divided into multiple groups.
In this heterogeneous setting, agents in different groups
have different characteristics, \eg, size and speed,
or play different roles in the task. In a heterogeneous setting, a permutation invariant critic is not directly applicable, because the relation between two heterogeneous agents differs from the relation between two homogeneous agents. For instance, the interaction between two fast-moving agents differs from the interaction between a fast-moving  and a slow-moving agent. However, in the aforementioned critic $Q_\text{PIC}$, relations between all agents are modeled equivalently. 

To address this concern, for the heterogeneous setting, we propose to add node attributes to the PIC. With node attributes, the PIC can distinguish agents from  different groups. Specifically, for group $j$, we construct a group attribute $g_j \in \mathbb{R}^{K_g}$, where $K_g$ denotes the dimension of the group attribute. Let $\boldsymbol{z}^t_i \in \mathbb{R}^{(K_a + K_o)}$ denote the input representation of agent $i$, \ie, the $i$-th row of $\boldsymbol{z}^t$. We obtain the augmented representation $\hat{\boldsymbol{z}}^t_i$ via
$
\hat{\boldsymbol{z}}^t_i  : = [\boldsymbol{z}^t_i,g_{G(i)}],
$
where $G(i)$ denotes the group index of agent $i$. We perform the augmentation for each agent to obtain the augmented representation $\hat{\boldsymbol{z}}^t \in \mathbb{R}^{N \times (K_a + K_o + K_g)}$. Using~\Eqref{eq:q_homo} and setting $K_\text{in}^{(1)} = K_a + K_o + K_g$ and $\bold h^{(0)}=\hat{\boldsymbol{z}}^t$,
results in a PIC that can handle environments with heterogeneous agents. 

\subsection{Improved MPE Environment}
The multiple particle environment (MPE)~\cite{maddpg, Mordatch17} is a multi-agent environment which contains a variety of tasks and provides a challenging open source platform for our community to evaluate new algorithms. However, the MPE targets only settings with a small number of agents. Specifically, we found it challenging to train more than 30 agents  as it takes more than 100 hours. 
To scale the MPE to more agents, we improve the implementation of the MPE. More specifically, we develop vectorized versions for many of the  computations, such as computing the force between agents, computing the collision between agents, \etc.
Moreover, for tasks with global rewards, instead of computing rewards for each agent, we only compute the reward once and send it to all agents. With this improved MPE, we can  train up to 200 agents within one day.

%% file: exp.tex

\section{Experiments}
\vspace{-0.2cm}
\label{sec:exp}
In this section, we first introduce tasks in the multiple particle environment (MPE)~\cite{maddpg, Mordatch17}. We then present the details of the experimental setup, evaluation protocol, and our results. 
\input{quan_results}

\noindent\textbf{Environment.} 
We evaluate the proposed approach on an improved version of MPE~\cite{maddpg, Mordatch17}. We consider the following four tasks:

\begin{itemize}[noitemsep,topsep=0pt,parsep=0pt,partopsep=0pt]
\item \emph{Cooperative navigation}: $N$ agents move cooperatively to cover $L$ landmarks in the environment. The reward encourages the agents to get close to landmarks. 
\item \emph{Prey and predator}: $N$ slower predators work together to chase $M$ fast-moving preys. The predators get positive reward when colliding with preys. {Preys are environment controlled.} 
\item \emph{Cooperative push}: $N$ agents work together to push a large ball to a landmark. The agents get rewarded when the large ball approaches the landmark. 
\item \emph{Heterogeneous navigation}: $N / 2$ small and fast agents and $N / 2$ big and slow agents work cooperatively to cover $N$ landmarks. The reward encourages the agents to get close to landmarks. If a small agent collides with a big agent, a large negative reward is received. 
\end{itemize}

\input{reward_fig}
\input{q_loss_fig}

Note \emph{cooperative navigation}, \emph{prey and predator} and \emph{cooperative push} are environments with homogeneous agents. 
In each task, the reward is global for cooperative agents, \ie, cooperative agents always receive the same reward in each timestep.

\noindent\textbf{Experimental Setup.} We use MADDPG~\cite{maddpg} with a classic MLP critic as the baseline. We implement MADDPG and the proposed approach in Pytorch~\cite{pytorch}.
To ensure the correctness of our implementation, we compare it with the official MADDPG  code~\cite{maddpg_code} on MPE. Our implementation reproduces the results of the official code. Please see~\tabref{tb:base} in the supplementary for the results.
Following MADDPG~\cite{maddpg}, the actor policy is parameterized by a two-layer MLP with 128 hidden units per layer, and ReLU  activation function. The MLP critic of the MADDPG baseline has the same architecture as the actor. Our permutation invariant critic (PIC) is a two-layer graph convolution net (GCN)  with 128 hidden units per layer and a max pooling at the top. The activation function for GCN is also a ReLU. Following MADDPG, the Adam optimizer is used. The learning rates for the actor, MLP critic, and our permutation invariant critic are $0.01$. The learning rate is linearly decreased to zero at the end of training. Agents are trained for $60,000$ episodes in all tasks (episode length is either $25$ or $50$ steps). The size of the replay buffer is one million, the batch size is $1024$, and the discounted factor $\gamma = 0.95$. In an eight-agent environment, the MLP critic and the PIC have around 44k and 40k trainable parameters respectively. In an 100-agent environment, the MLP critic has 413k trainable parameters while the PIC has only 46k parameters.

\noindent\textbf{Evaluation Protocol.} To ensure  a fair and rigorous evaluation, we follow the strict evaluation protocols suggested by~\citet{Colas18} and~\citet{Henderson17}. For each experiment, we report \emph{final metrics} and \emph{absolute metrics}. 
The final metric~\cite{Colas18} is the average reward over the last $10,000$ evaluation episodes, \ie, $1,000$ episodes for each of the last ten policies during training. The absolute metric~\cite{Colas18} is the best policy's average reward over $1,000$ evaluation episodes. To analyze the significance of the reported improvement over the baseline, we perform a two-sample t-test and boostrapped estimation of the $95\%$ confidence interval for the mean reward difference obtained by the baseline and our approach. We use the Scipy~\cite{scipy} implementation for the t-test, and the Facebook Boostrapped implementation with $10,000$ boostrap samples for confidence interval estimation.
All  experiments are repeated for five runs with different random seeds. 

\input{t_test_results}
\noindent\textbf{Results.} 
We compare the proposed permutation invariant critic (PIC) with an MLP critic and  an MLP critic with data augmentation.  Data augmentation shuffles the order of agents' observations and actions when training the MLP critic, which is considered to be a simple way to alleviate the ordering issue. The results are summarized in~\tabref{tb:main}, where $N$ denotes the number of agents, `final' and `absolute' are the final metric and absolute metric respectively.  We observe that data augmentation does not boost the performance of an MLP critic much. In some cases, it even deteriorates the results.  In contrast, as shown in~\tabref{tb:main}, a permutation invariant critic outperforms the MLP critic and the MLP critic with data augmentation by about $15 \% - 50 \%$ in all tasks.

The training curves are given in \figref{fig:r_plot} and \figref{fig:q_plot}.
As shown in \figref{fig:q_plot}, the loss (\Eqref{eq:maddpg_q_loss}) of the permutation invariant critic is  lower than that of MLP critics.
Please see the supplementary material for more training curves. \figref{fig:q_loss_and_time} (top)
shows the ratio of the MLP critic's average loss to the permutation invariant critic's loss on cooperative navigation environments. We observed that the ratio grows when the number of agents increases. This implies that a permutation invariant approach gives much more accurate value estimation than the MLP critic particularly when the number of agents is large. 

To confirm that the performance gain of our permutation invariant critic is significant, we report the 2-sample t-test results and the $95 \%$ boostrapped confidence interval on the mean difference of our approach and the baseline MADDPG with MLP critic. For the 2-sample t-test, the t-statistic and the p-value are reported. The difference is considered significant when the p-value is smaller than $0.05$. \tabref{tb:t_test} 
summarizes the analysis on cooperative navigation environments.  As shown in~\tabref{tb:t_test}, 
 the p-value is much smaller than $0.05$ which suggests the improvement is significant. In addition, positive confidence interval and means suggest that our approach achieves higher rewards than the baseline. Please see the supplementary material for additional analysis.

{In addition to fully connected graphs we also tested $K$-nearest neighbor graphs, \ie, each node is connected only to its $K$ nearest neighbors. The distance between two nodes is the physical distance between the corresponding agents in the environment. We found that using fully connected graphs achieves better results than using $K$-nearest neighbor graphs.  We report absolute rewards for ($K=2$, $K=6$, fully connected graph): the results on  cooperative navigation ($N=15$), prey and predator ($N=15$), cooperative push ($N=15$), and heterogeneous navigation ($N=16$) are $(-2162, -2059, -1977)$, $(3014, 3163, 10239)$, $(-2370, -2270, -2225)$, and $(-2635, -2552, -1820)$ respectively.
}

\figref{fig:q_loss_and_time} (bottom) 
compares training time of MADDPG on the original MPE and our improved MPE. MADDPG is trained for $60,000$ episodes with a PIC. As shown in~\figref{fig:q_loss_and_time} (bottom),
training $30$ agents in the original MPE environment takes more than 100 hours. In contrast, with the improved MPE environment, we can train 30 agents within five hours, and scale to 200 agents within a day of training. 

%% file: quan_results.tex
\begin{table}[t] 
\vspace{-1cm}
\begin{center}
\small
\scalebox{1}{
{
\begin{tabular}{c|c|cccccc}
\specialrule{.15em}{.05em}{.05em} 
                      &  & \multicolumn{2}{c}{\makecell{MLP Critic}} & \multicolumn{2}{c}{\makecell{MLP Critic \\ + Data Augaumentation} }  & \multicolumn{2}{c}{\makecell{Ours \\ Permutation Invariant \\ Critic} } \\ \toprule\toprule
                      				&\#  of agents         & final             & absolute             & final              & absolute           & final            & absolute          \\ \hline

\multirow{6}{*}{\makecell{Cooperative \\ Navigation}}     
& N=3    &         -362.73 &	-362.71	& -361.76 &	-361.55 &	\bf{-355.99} & \bf{-355.74}               \\
 &N=6      &         -3943.2 & 	-3933.3 & -4025.4 & -4016.2            &      \bf{-3383.2}          &     \bf{-3381.8}            \\
 & N=15  		 &	-6489.7 & -6394.3 & -6280.0 & -6222.4 & \bf{-1999.1} &	 \bf{-1977.6}              \\
 &N=30			  &	-20722 &	-20583 &	-21205 &	-20840 &	\bf{-11363} &	\bf{-11294}              \\ 
&N=100  &         	-128100 & -128086 & -128024 & -128013 & \bf{-71495} & \bf{-71074}             \\
 & N=200 		&	-502349 &  -502348 & -509963 & -507457 & \bf{-436215} & \bf{-433846}                 \\

\midrule

\multirow{5}{*}{\makecell{Prey  \\ \& \\ Predator}}     
& N=3    &        38.52  & 40.91 & 43.78 & 44.99 & \bf{65.16} & \bf{67.69}              \\
 &N=6      &         26.15 &	30.34	& -24.84 & -17.49 & \bf{176.70} & \bf{184.22}           \\
 & N=15  		 &	3982.80 & 4416.23 & 4198.41 & 4401.64 & \bf{10139} & \bf{10239}              \\
 &N=30			  &	-377.19	& -386.49	& -93.86 &	-75.59 & \bf{6662.1}	& \bf{6745.6}               \\ 
&N=100 &         	19894 &	21114 &	28741 & 29347 &	\bf{99391} &	\bf{100812}              \\
\midrule

\multirow{5}{*}{\makecell{Cooperative \\ Push}}     
& N=3    &         -171.26 &	-170.20 & -171.96 & -171.74 & \bf{-155.17} & \bf{-155.10}               \\
 &N=6      &         -561.73 & -542.29 & -672.80 & -672.27 & \bf{-401.79} & \bf{-395.70}           \\
 & N=15  		 &	-2538.3 & -2536.0 & 	-2645.2 & 	-2610.1 & 	\bf{-2231.1} & \bf{-2225.7}                \\
 &N=30 			  &	-3499.7 & -3465.9 &	-3761.5 &	-3688.3 &	\bf{-3117.1}& \bf{-3094.3}              \\

\midrule
\multirow{5}{*}{\makecell{Heterogeneous \\ Navigation}}     
& N=4    &         -100.74 &	-100.35 &	-98.96 & -98.67 & \bf{-83.84} & \bf{-83.54}              \\
 &N=8      &         -398.31 & -397.66 & -683.24 & -684.39 & \bf{-398.31} & \bf{-397.66}           \\
 & N=16  		 &	-3410.0 & -3405.9 &	-3479.0 &	-3470.8 &	\bf{-1825.8} &	\bf{-1820.4}          \\
 &N=30 		&	-12944 &	-12934 &	-12812 &	-12809 &	\bf{-6293} &	\bf{-6270 }            \\ 
&N=100  &         	-121563 &  -121472 & -130028 & -129436 & \bf{-94324} &  \bf{-93996}           \\
\specialrule{.15em}{.05em}{.05em} 
\end{tabular}
}}
\caption{Average rewards of our approach and two MADDPG baselines.  `final' represents final metric~\cite{Colas18}, which is the average reward over the last 10,000 evaluation episodes, \ie, 1000 episodes for each of the last ten policies during training. `absolute' represents absolute metric~\cite{Colas18}, which is the best policy's average reward over 1,000 evaluation episodes. 
}
\label{tb:main}
\end{center}\vspace{-0.7cm}
\end{table}

%% file: reward_fig.tex
\begin{figure*}[t]
\vspace{-0.5cm}
\centering
\begin{tabular}{ccc}
\includegraphics[width=0.33\textwidth]{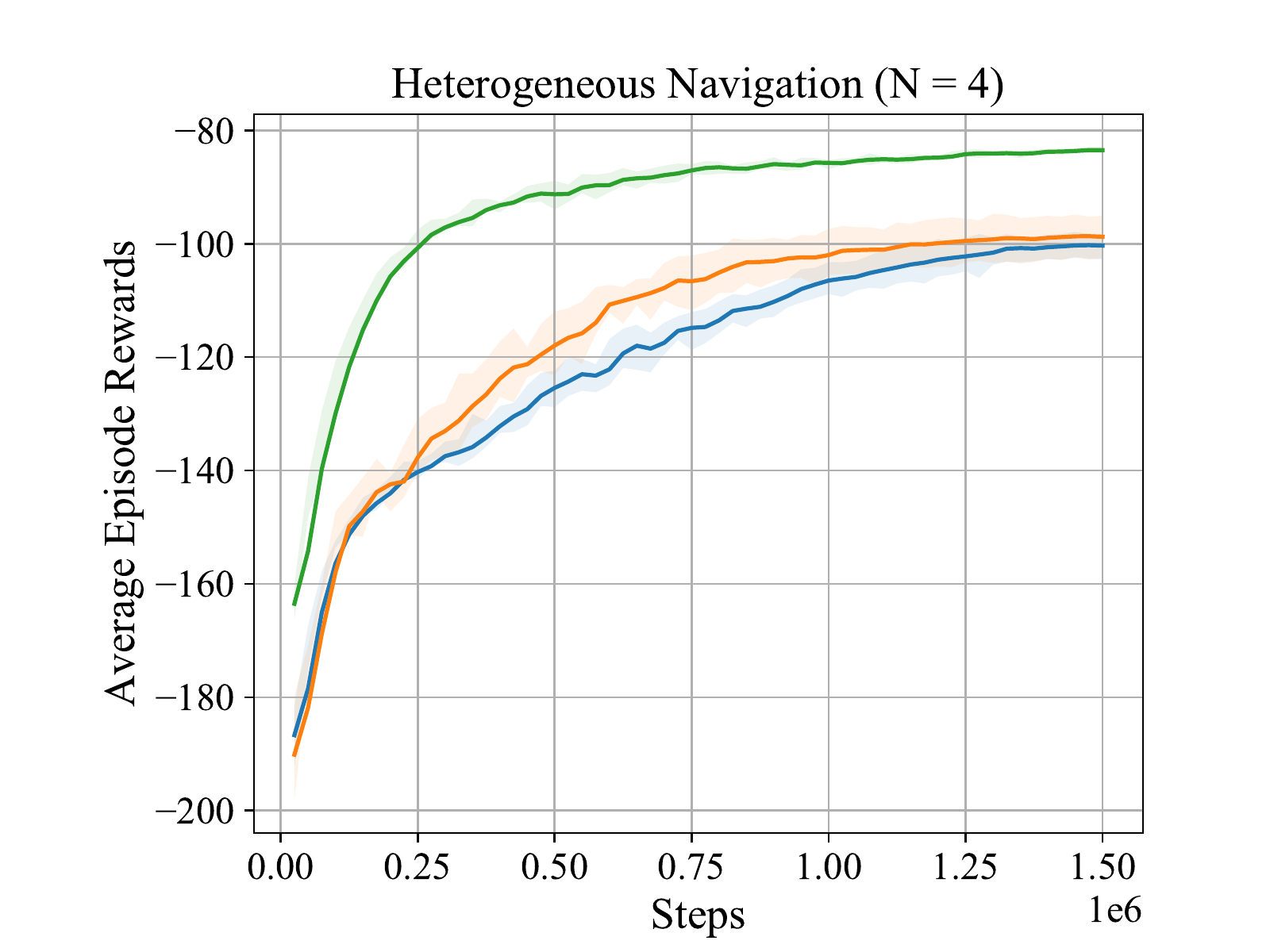}
&
\hspace{-0.6cm}\includegraphics[width=0.33\textwidth]{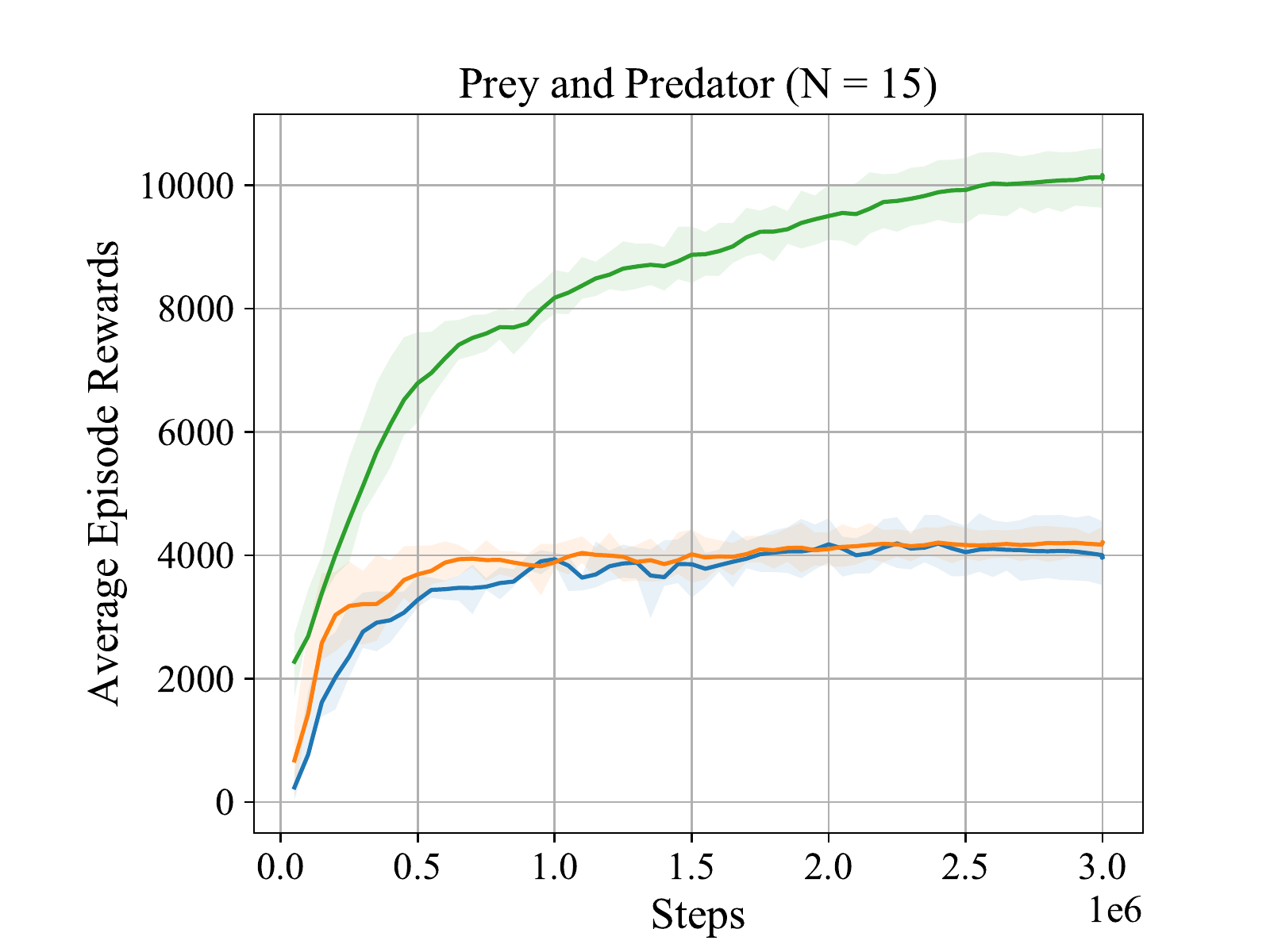}
&
\hspace{-0.6cm}\includegraphics[width=0.33\textwidth]{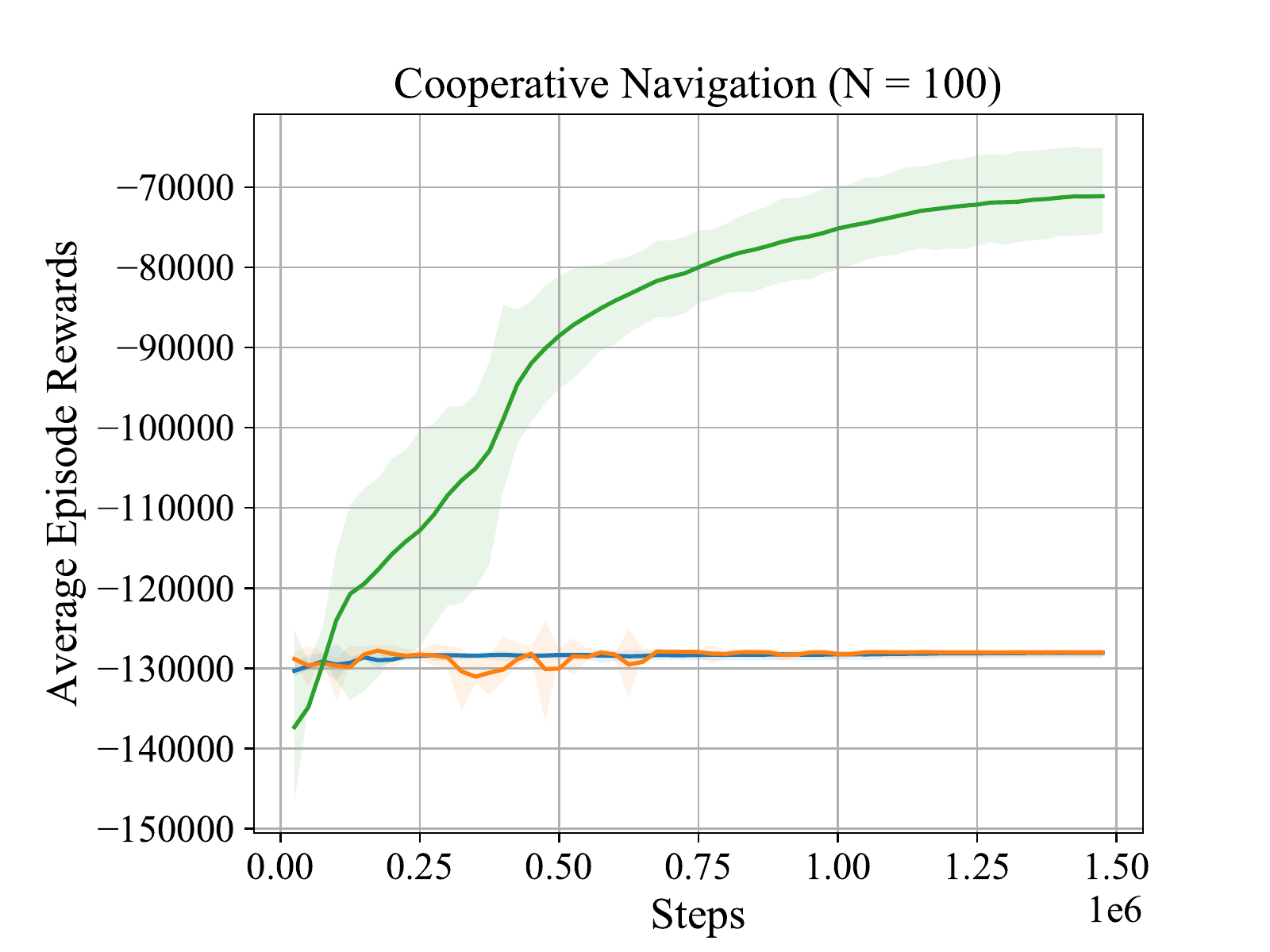}\\
\multicolumn{3}{c}{
\includegraphics[width=0.9\textwidth]{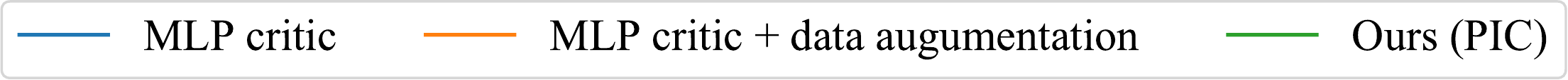}}\\
\end{tabular}
\vspace{-0.3cm}
\caption{Comparison of average episode reward.
}
\label{fig:r_plot}
\vspace{-0.4cm}
\end{figure*}

%% file: q_loss_fig.tex
\begin{figure*}[t]
\centering
\begin{tabular}{ccc}

\includegraphics[width=0.33\textwidth]{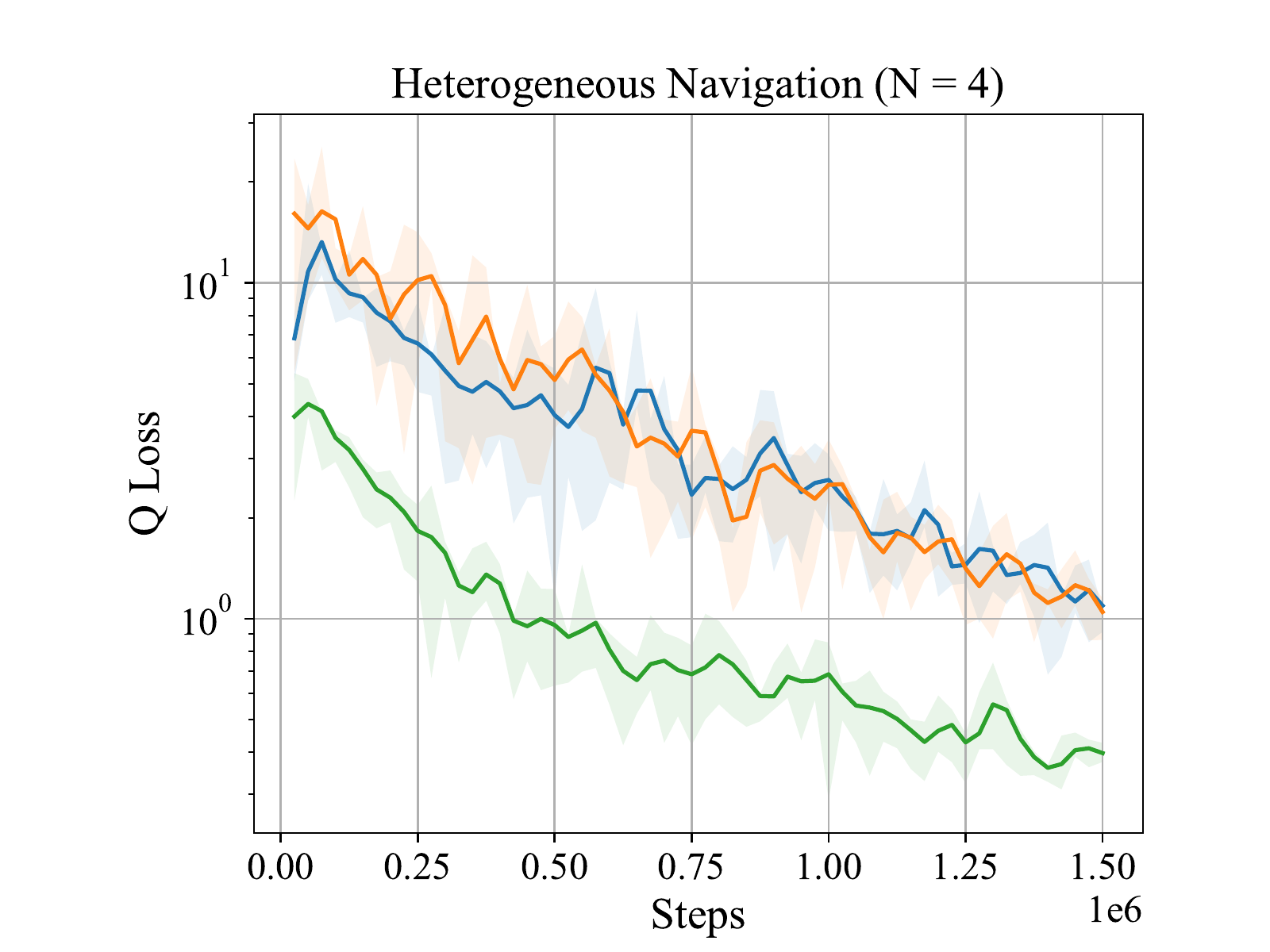}
&
\hspace{-0.6cm}
\includegraphics[width=0.33\textwidth]{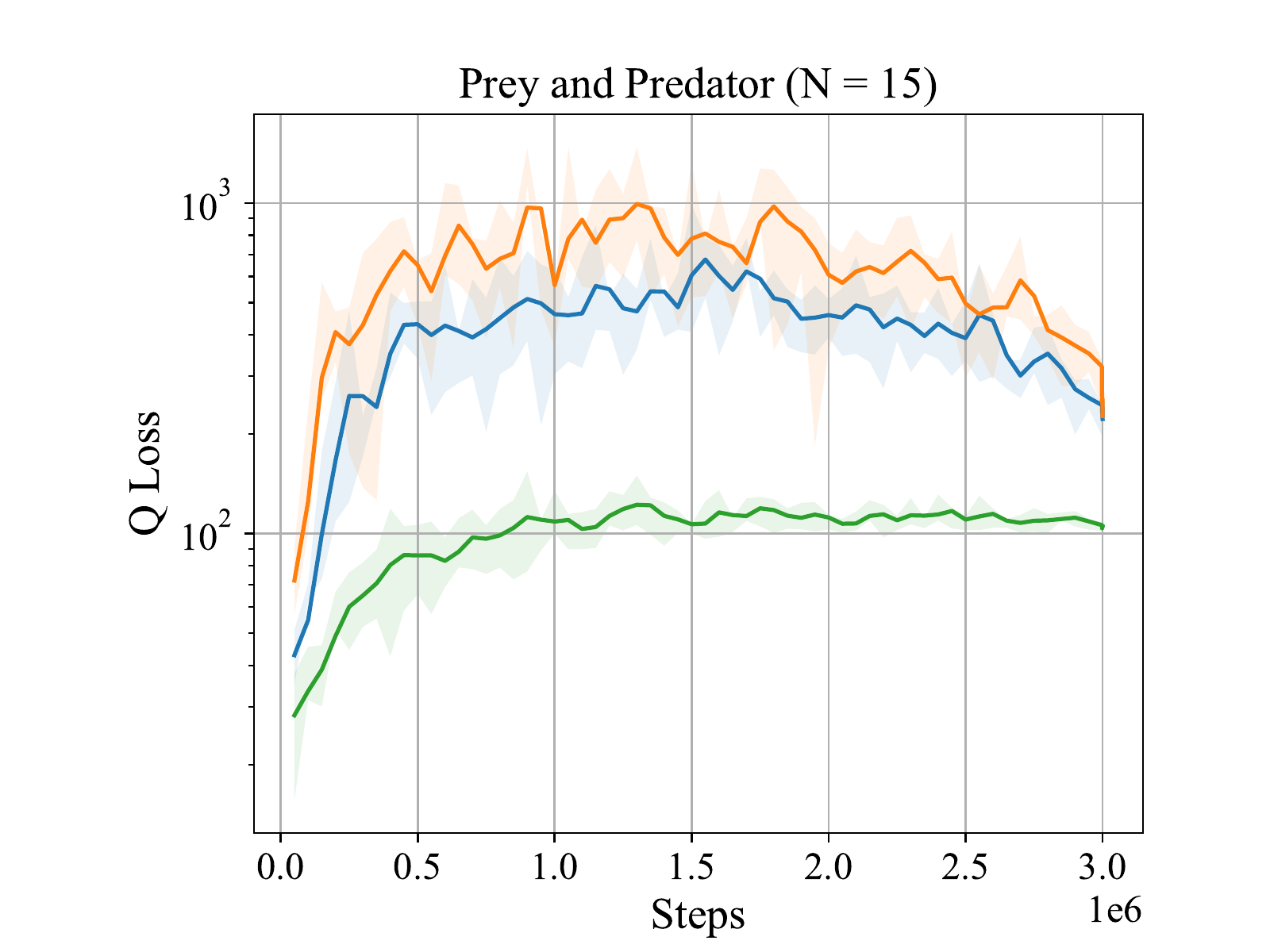}
&
\hspace{-0.6cm}
\includegraphics[width=0.33\textwidth]{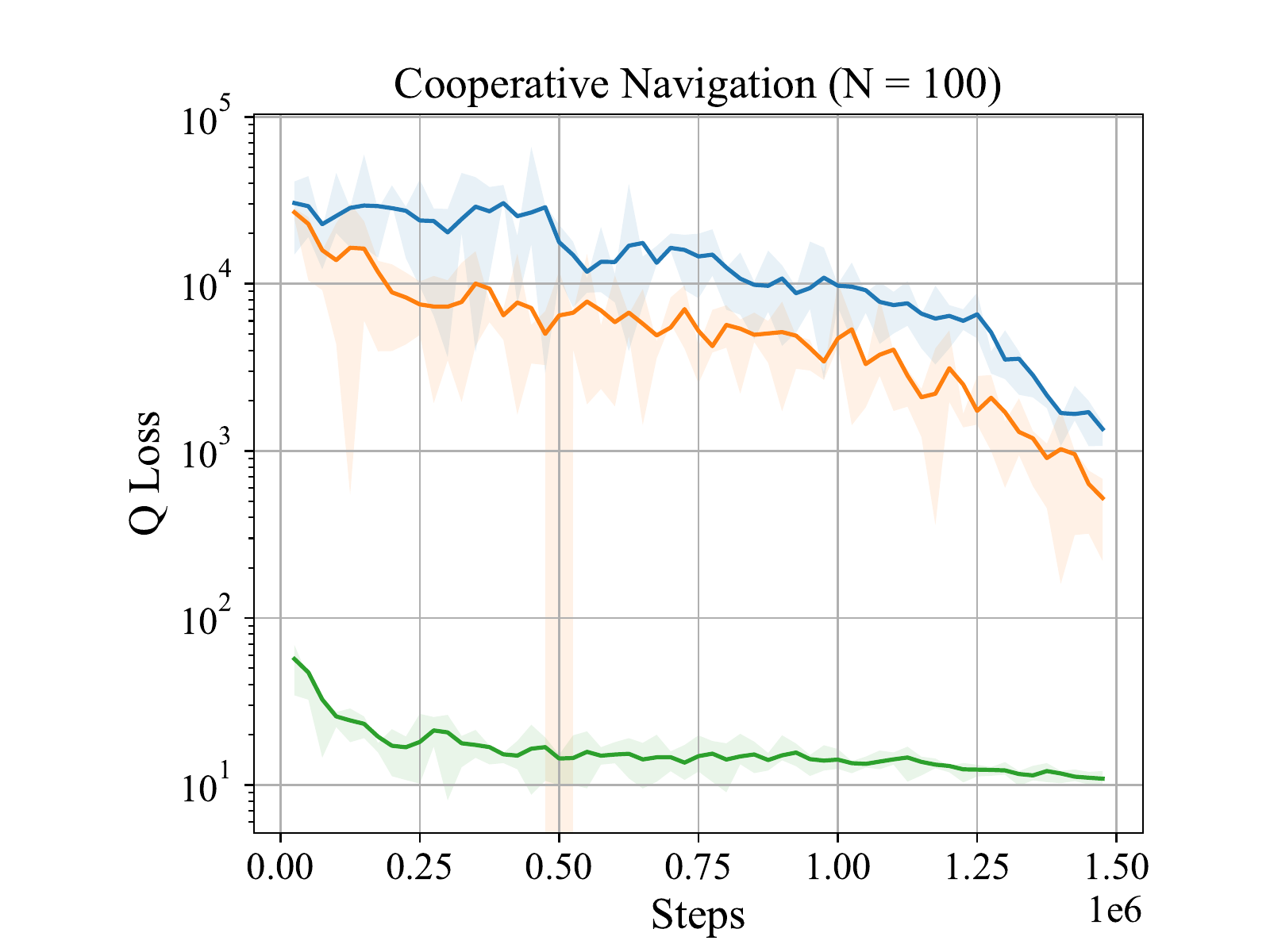}\\
\multicolumn{3}{c}{
\includegraphics[width=0.9\textwidth]{fig/legend.pdf}}\\
\end{tabular}%
\vspace{-0.3cm}%
\caption{Comparison of average Q loss (\Eqref{eq:maddpg_q_loss}).
}
\label{fig:q_plot}
\vspace{-0.8cm}
\end{figure*}

%% file: t_test_results.tex
\begin{figure*}
\vspace{-0.9cm}
\begin{minipage}{.62\textwidth}
      \centering
\begin{tabular}{c|c|cc}
\specialrule{.15em}{.05em}{.05em} 
                                       & \multicolumn{1}{c}{}      &  \multicolumn{1}{c}{{t-test}}               &  \multicolumn{1}{c}{{Boostrap C.I.}}        \\ \toprule\toprule
                                       && t-stat.	(p-value)&	mean (C.I.) \\\midrule
\multirow{2}{*}{\makecell{N=3}}     	& \multicolumn{1}{c|}{abs.}  &   1.42 (1.9e{-1})      &  5.81 (-1.47, 13.09)                  \\
                                      		& \multicolumn{1}{c|}{final}  &   1.41  (1.9e{-1})      &  5.76 (-1.65, 13.18)       \\ \midrule
\multirow{2}{*}{\makecell{N=6}}       	& \multicolumn{1}{c|}{abs.}  &   9.46 (3.2e{-5})     &         634.4 (487.9, 757.1)                   \\
                                       		& \multicolumn{1}{c|}{final} &           9.34 (4.1e{-5})            &     642.2 (496.1, 762.3)            \\ \midrule
\multirow{2}{*}{\makecell{N=15}}         	& \multicolumn{1}{c|}{abs.} &  21.04 (2.5e{-5})      & 4244 (3819, 4575)                           \\
                                       		& \multicolumn{1}{c|}{final} &  20.6 (2.8e{-5})      & 4280 (3854, 4616)                 \\\midrule
\multirow{2}{*}{\makecell{N=30}}  		& \multicolumn{1}{c|}{abs.} &  3.17 (2.2e{-2})    	&  9546    (6719, 12639)                         \\
                                       		& \multicolumn{1}{c|}{final} &  3.10 (2.1e{-2})       & 9841    (6692, 13201)             \\\midrule
\multirow{2}{*}{\makecell{N=100}} 		& \multicolumn{1}{c|}{abs.} & 17.8 (2.9e{-3})        & 56939    (51842, 62764)                            \\
                                        		& \multicolumn{1}{c|}{final} & 17.3 (3.0e{-3})        & 56529   (51153, 62343)             \\\midrule
\multirow{2}{*}{N=200}           & \multicolumn{1}{c|}{abs.} & 2.72 (4.9e{-2})        & 73611    (25930, 116198)                           \\
                                       		& \multicolumn{1}{c|}{final} & 2.66 (4.8e{-2})        &  73747    (26557, 116457)         \\
\specialrule{.15em}{.05em}{.05em} 
\end{tabular}
\captionof{table}{T-test and boostrap confidence interval of mean difference between MLP critic and our permutation invariant critic on cooperative navigation.
}\label{tb:t_test}
\end{minipage} 
\hspace{0.01\textwidth}
\begin{minipage}{0.35\textwidth}
      \centering
   \begin{minipage}{\textwidth}
    \centering
   	\includegraphics[height=4.cm, trim={1cm 0 1cm .1cm},clip]{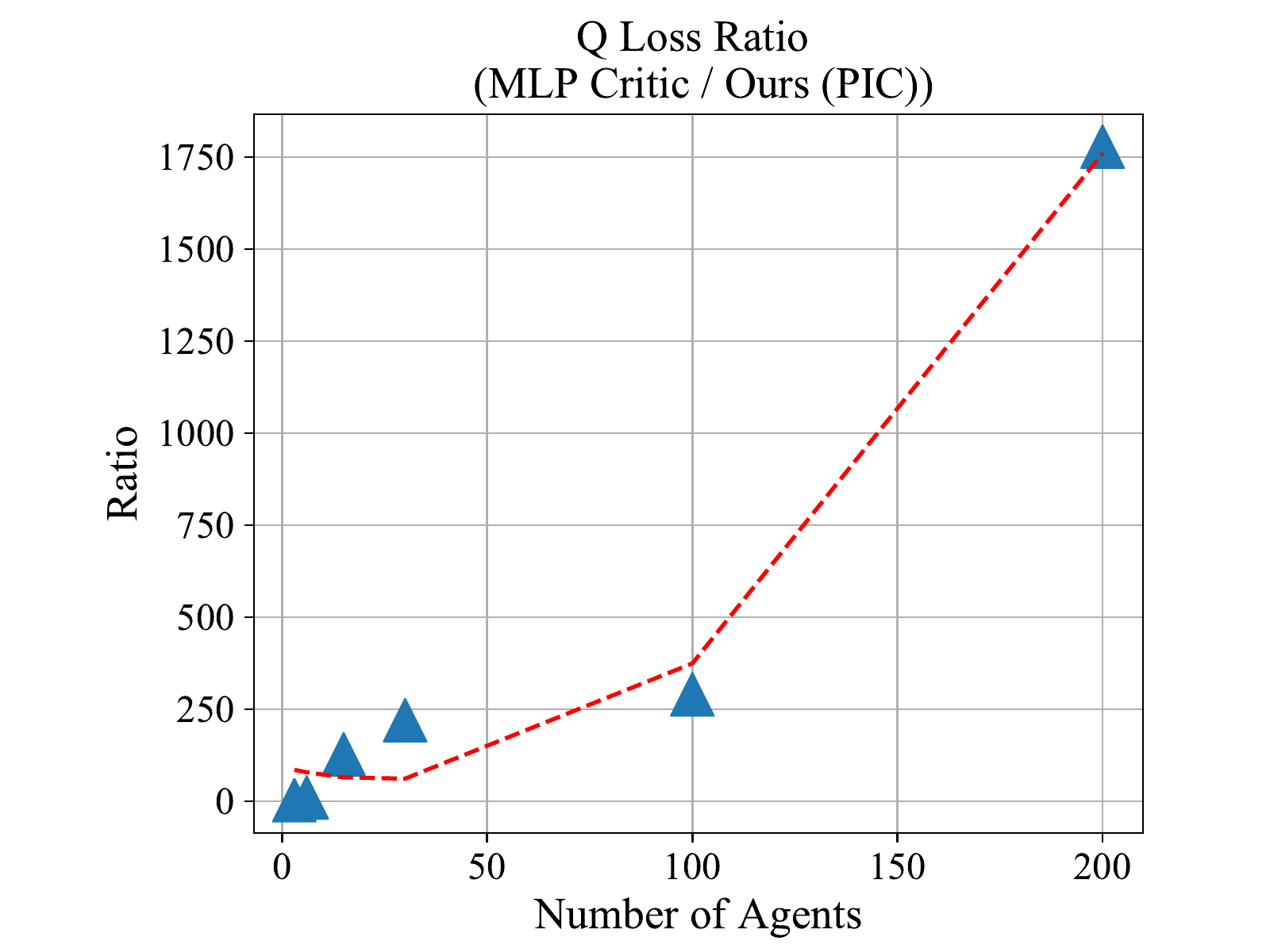}
   \end{minipage} 
      \begin{minipage}{\textwidth}
       \centering
   	\includegraphics[height=4.cm, trim={1cm .15cm 1cm .1cm},clip]{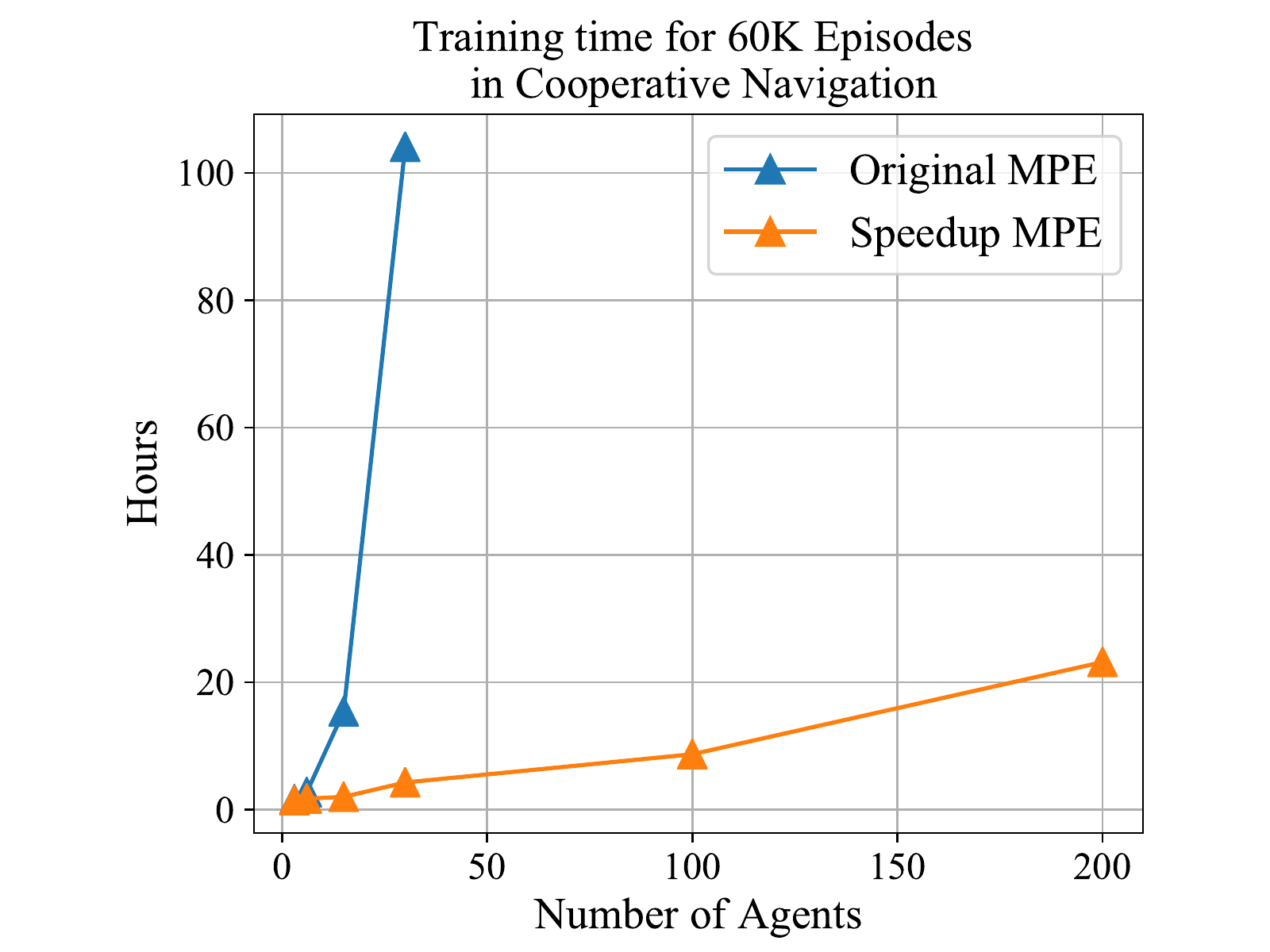}
   \end{minipage} 
    \centering
    \vspace{-0.2cm}
\captionof{figure}{\textbf{Top}: Average critic loss ratio (MLP / Ours). \textbf{Bottom:} Training time comparison.}    \label{fig:q_loss_and_time}
\end{minipage}
\vspace{-0.8cm}
\end{figure*}

%% file: conc.tex
\vspace{-0.4cm}
\section{Conclusion}
\vspace{-0.3cm}
We propose and study permutation invariant critics to estimate a consistent value for all permutations of the agents' order in multi-agent reinforcement learning. Empirically, we demonstrate that a permutation invariant critic outperforms classical deep nets on a variety of multi-agent environments, both homogeneous and heterogeneous. Permutation invariant critics lead to better sample complexity and permit to scale learning to environments with a large number of agents. We think that the permutation invariance property is important for deep multi-agent reinforcement learning. Going forward we will study other designs to scale to an even larger number of agents.

%% file: ack.tex
\noindent\textbf{Acknowledgements:} 
This work is supported in part by NSF under Grant $\#$1718221 and MRI $\#$1725729, UIUC, Samsung, 3M, Cisco Systems Inc.\ (Gift Award CG 1377144), Adobe, and a Google PhD Fellowship to RY. We thank NVIDIA for providing GPUs used for this work and Cisco for access to the Arcetri cluster. 

%% file: supp.tex

\appendix
\renewcommand{\thetable}{A\arabic{table}}
\setcounter{table}{0}
\setcounter{figure}{0}
\renewcommand{\thetable}{A\arabic{table}}
\renewcommand\thefigure{A\arabic{figure}}
\renewcommand{\theHtable}{Appendix.\thetable}
\renewcommand{\theHfigure}{Appendix.\thefigure}

\newcommand*{\dictchar}[1]{
    \clearpage
    \twocolumn[
    \centerline{\parbox[c][3cm][c]{\textwidth}{
            \centering
            \fontsize{14}{14}
            \selectfont
            {#1}}}]
}

{\centering \Large \textbf{Supplementary material: PIC: Permutation Invariant Critic for Multi-agent Deep Reinforcement Learning}}

 \section{MADDPG Baseline}
\label{sec:appendix1}
We implement the MADDPG baseline in Pytorch. To ensure we implement MADDPG correctly, we compare the performance of our implementation with  MADDPG's official code on the MPE. As shown in~\tabref{tb:base}, our implementation reproduces the results of MADDPG  in all environments.

 \begin{table*}[b]
\centering
\caption{Comparison of the official MADDPG code~\cite{maddpg_code} and our MADDPG implementation on MPE. 
}
\hspace*{-1cm}
\setlength{\tabcolsep}{3pt}
{\footnotesize
\begin{tabular}{c|ccccc} \toprule
                     & \makecell{Cooperative \\Navigation} &  \multicolumn{2}{c}{Prey and Predator} & \multicolumn{2}{c}{Push}  \\ \midrule\midrule
                     & good                           & adversary            & good         & adversary       & good         \\ \hline
MADDPG official code & -379.57                  &    2.91            & 5.48                 & -7.05        & -1.67                    \\
Our implementation   & -367.75                   & 5.97            & 14.79                & -6.54        & -1.81                     \\ \bottomrule
\end{tabular}
}
\label{tb:base}
\end{table*}
 
\section{Environment Details and Group Embedding}
\label{sec:appendix2}
In this section, we first provide details of observation and action space in each environment we considered in our experiments. Subsequently, we discuss a PIC's group embedding. 

In all the four environments, the action dimension is five. One dimension is no-op. The other four dimensions represent the left, right, forward,  and backward force applied to the particle agent. 
An agent's observation  always contains its location and velocity. Depending on the environment the observation may contain relative location and velocity of neighboring agents and landmarks. In our experiments, the number of visible neighbors in an agent's observation is equal to or less than ten because we empirically found a larger number of visible neighbors to not boost the performance of MADDPG but rather slow  the training speed. {Note, the number of visible neighbors in an agent's observation is different from the $K$-nearest neighbor graph discussed in \secref{sec:exp}. The $K$ in the $K$-nearest neighbor graph refers to the number of agent observations and actions which are used as input to the centralized critic, while the number of visible neighbors in an agent's observation is a characteristic of an environment. 
The details of the observation representation for each environment are as follows: 

\begin{itemize}[noitemsep,topsep=0pt,parsep=0pt,partopsep=0pt]
\item \emph{Cooperative navigation}: An agent observes its location and velocity, and the relative location of the nearest $k$ landmarks and agents. $k$ is 2, 5, 5, 5, 5, 5 for an environment with 3, 6, 15, 30, 100, 200 agents. As a result, the observation dimension is 14, 26, 26, 26, 26, 26 for an environment with 3, 6, 15, 30, 100, 200 agents. 

\item \emph{Prey and predator}: A predator observes its location and velocity, the relative location of the nearest $l$ landmarks and fellow predators, the relative location and velocity of the $k$ nearest preys. $(k, l)$ is (2, 1), (3, 2), (5, 5),  (5, 5) (5, 5) for an environment with 3, 6, 15, 30, 100 agents. As a result, the observation dimension is 16, 28, 34, 34, 34 for an environment with 3, 6, 15, 30, 100 agents. 

\item \emph{Cooperative push}:  An agent observes its location and velocity,  the relative location of the target landmark and the large ball,  and the relative location of the $k$ nearest agents. $k$ is 2, 5, 10, 10 for an environment with 3, 6, 15, 30 agents. As a result, the observation dimension is 12, 18, 28, 28 for an environment with 3, 6, 15, 30 agents. 

\item \emph{Heterogeneous navigation}: An agent observes its location and velocity, and the relative location of the nearest $k$ landmarks and agents. $k$ is 2, 5, 5, 5, 5  for an environment with 3, 6, 15, 30, 100 agents. Consequently, the observation dimension is 14, 26, 26, 26, 26 for an environment with 3, 6, 15, 30, 100 agents.
\end{itemize}
In the heterogeneous environment \emph{Heterogeneous navigation}, the number of groups is two, \ie,  two groups of agents that have different characteristics. 
For the PIC, the group embedding for each group is a two-dimensional vector, \ie, $g_j \in \mathbb{R}^{2}$ for each group $j \in \{1, 2\}$ in the environment. 
We train the embedding along with the network parameters. The group embedding is randomly initialized from a normal distribution $\mathcal {N}(0, \text{diag}(1)_2)$.

\section{T-test and Training Curves}\label{sec:curves}
The two-sample t-test and the $95\%$ confidence interval of the mean difference of the MLP critic and the PIC are summarized in~\tabref{tb:t_test_spreada},~\tabref{tb:t_test_taga},~\tabref{tb:t_test_pusha}, and~\tabref{tb:t_test_heteroa}. p-values smaller than $0.05$ and positive confidence intervals indicated that PIC's improvement over the MLP critic is significant.
The training curves for MADDPG with MLP critic and our PIC are shown in~\figref{fig:r_plot_all1} and~\figref{fig:r_plot_all2}. The PIC outperforms the MLP critic in all the environment settings.

\begin{table}[]

\begin{tabular}[t]{cc}
\adjustbox{valign=t}{
    \begin{minipage}{.5\linewidth}
      \centering

       \caption{ Ours (PIC)  \vs MLP critic in the cooperative navigation environment.}

\setlength{\tabcolsep}{2pt}
\scriptsize
\begin{tabular}{c|c|cc}
\specialrule{.15em}{.05em}{.05em} 
                                       & \multicolumn{1}{c}{}      &  \multicolumn{1}{c}{{t-test}}               &  \multicolumn{1}{c}{{Boostrap C.I.}}        \\ \toprule\toprule
                                       && t-stat.	(p-value)&	mean (C.I.) \\\midrule
\multirow{2}{*}{\makecell{N=3}}     	& \multicolumn{1}{c|}{abs.}  &   1.42 (1.9e{-1})      &  5.81 (-1.47, 13.09)                  \\
                                      		& \multicolumn{1}{c|}{final}  &   1.41  (1.9e{-1})      &  5.76 (-1.65, 13.18)       \\ \midrule
\multirow{2}{*}{\makecell{N=6}}       	& \multicolumn{1}{c|}{abs.}  &   9.46 (3.2e{-5})     &         634.4 (487.9, 757.1)                   \\
                                       		& \multicolumn{1}{c|}{final} &           9.34 (4.1e{-5})            &     642.2 (496.1, 762.3)            \\ \midrule
\multirow{2}{*}{\makecell{N=15}}         	& \multicolumn{1}{c|}{abs.} &  21.04 (2.5e{-5})      & 4244 (3819, 4575)                           \\
                                       		& \multicolumn{1}{c|}{final} &  20.6 (2.8e{-5})      & 4280 (3854, 4616)                 \\\midrule
\multirow{2}{*}{\makecell{N=30}}  		& \multicolumn{1}{c|}{abs.} &  3.17 (2.2e{-2})    	&  9546    (6719, 12639)                         \\
                                       		& \multicolumn{1}{c|}{final} &  3.10 (2.1e{-2})       & 9841    (6692, 13201)             \\\midrule
\multirow{2}{*}{\makecell{N=100}} 		& \multicolumn{1}{c|}{abs.} & 17.8 (2.9e{-3})        & 56939    (51842, 62764)                            \\
                                        		& \multicolumn{1}{c|}{final} & 17.3 (3.0e{-3})        & 56529   (51153, 62343)             \\\midrule
\multirow{2}{*}{N=200}           & \multicolumn{1}{c|}{abs.} & 2.72 (4.9e{-2})        & 73611    (25930, 116198)                           \\
                                       		& \multicolumn{1}{c|}{final} & 2.66 (4.8e{-2})        &  73747    (26557, 116457)         \\
\specialrule{.15em}{.05em}{.05em} 
\end{tabular}

\label{tb:t_test_spreada}    \end{minipage}} &
\adjustbox{valign=t}{
    \begin{minipage}{.5\linewidth}
      \centering
      \caption{ Ours (PIC) \vs MLP critic in the prey and predator environment.}
   
\setlength{\tabcolsep}{2pt}
\scriptsize
\begin{tabular}{c|c|cc}
\specialrule{.15em}{.05em}{.05em} 
                                       & \multicolumn{1}{c}{}      &  \multicolumn{1}{c}{{t-test}}               &  \multicolumn{1}{c}{{Boostrap C.I.}}        \\ \toprule\toprule
                                       && t-stat.	(p-value)&	mean (C.I.) \\\midrule
\multirow{2}{*}{\makecell{N=3}}     	& \multicolumn{1}{c|}{abs.}  &   6.61 (1.6e{-4})      &  22.7    (15.2, 29.7)                  \\
                                      		& \multicolumn{1}{c|}{final}  &   5.38  (8.0e{-4})      &  21.3    (13.6, 29.6)      \\ \midrule
\multirow{2}{*}{\makecell{N=6}}       	& \multicolumn{1}{c|}{abs.}  &   3.14 (3.4e{-2})     &         201    (88.2, 316)                  \\
                                       		& \multicolumn{1}{c|}{final} &           3.11 (3.5e{-2})            &     201    (86.4, 318)            \\ \midrule
\multirow{2}{*}{\makecell{N=15}}         	& \multicolumn{1}{c|}{abs.} &  20.34 (1.7e{-6})      & 5838    (5468, 6248)                        \\
                                       		& \multicolumn{1}{c|}{final} &  20.29 (5.7e{-7})      & 5941    (5609, 6303)                 \\\midrule
\multirow{2}{*}{\makecell{N=30}}  		& \multicolumn{1}{c|}{abs.} &  5.91 (4.0e{-3})    	&  6821    (5176, 9006)                         \\
                                       		& \multicolumn{1}{c|}{final} &  5.82 (4.3e{-3})       &  6755    (5106, 8925)             \\\midrule
\multirow{2}{*}{\makecell{N=100}} 		& \multicolumn{1}{c|}{abs.} & 6.92 (7.3e{-2})        & 92753    (82974, 102532)                           \\
                                        		& \multicolumn{1}{c|}{final} & 7.44 (7.1e{-2})        & 89980    (80796, 99163)            \\
\specialrule{.15em}{.05em}{.05em} 
\end{tabular}
\label{tb:t_test_taga}    \end{minipage} }\\

\adjustbox{valign=t}{
\begin{minipage}{.5\linewidth}
      \centering
      \caption{ Ours (PIC)  \vs MLP critic in the cooperative push environment.}
   
\setlength{\tabcolsep}{2pt}
\scriptsize
\begin{tabular}{c|c|cc}
\specialrule{.15em}{.05em}{.05em} 
                                       & \multicolumn{1}{c}{}      &  \multicolumn{1}{c}{{t-test}}               &  \multicolumn{1}{c}{{Boostrap C.I.}}        \\ \toprule\toprule
                                       && t-stat.	(p-value)&	mean (C.I.) \\\midrule
\multirow{2}{*}{\makecell{N=3}}     	& \multicolumn{1}{c|}{abs.}  &   4.03 (1.4e{-2})      &  16.6    (10.3, 24.4)                 \\
                                      		& \multicolumn{1}{c|}{final}  &   4.04  (1.4e{-2})      &  16.7    (10.4, 24.6)      \\ \midrule
\multirow{2}{*}{\makecell{N=6}}       	& \multicolumn{1}{c|}{abs.}  &   66.4 (1.0e{-10})     &          276.5   (269, 287)                   \\
                                       		& \multicolumn{1}{c|}{final} &           67.2 (1.3e{-11})            &     271.0   (263, 281)      \\ \midrule
\multirow{2}{*}{\makecell{N=15}}         	& \multicolumn{1}{c|}{abs.} &  4.90 (6.3e{-3})      & 384    (250, 507)                          \\
                                       		& \multicolumn{1}{c|}{final} &  5.12 (5.2e{-3})      & 414    (280, 546)                 \\\midrule
\multirow{2}{*}{\makecell{N=30}}  		& \multicolumn{1}{c|}{abs.} &  4.41 (7.1e{-3})    	&  593    (400, 778.316)                        \\
                                       		& \multicolumn{1}{c|}{final} &  4.64 (5.4e{-3})       &  644    (419, 844)            \\
\specialrule{.15em}{.05em}{.05em} 
\end{tabular}
\label{tb:t_test_pusha}    \end{minipage}} &
\adjustbox{valign=t}{
\begin{minipage}{.5\linewidth}
      \centering
       \caption{ Ours (PIC) \vs MLP critic in the heterogeneous navigation environment.}
   
\setlength{\tabcolsep}{2pt}
\scriptsize
\begin{tabular}{c|c|cc}
\specialrule{.15em}{.05em}{.05em} 
                                       & \multicolumn{1}{c}{}      &  \multicolumn{1}{c}{{t-test}}               &  \multicolumn{1}{c}{{Boostrap C.I.}}        \\ \toprule\toprule
                                       && t-stat.	(p-value)&	mean (C.I.) \\\midrule
\multirow{2}{*}{\makecell{N=3}}     	& \multicolumn{1}{c|}{abs.}  &   7.07 (1.7e{-3})      &  15.1    (11.8, 18.4)                 \\
                                      		& \multicolumn{1}{c|}{final}  &   6.76  (2.2e{-3})      &  15.1  (11.6, 18.6)       \\ \midrule
\multirow{2}{*}{\makecell{N=6}}       	& \multicolumn{1}{c|}{abs.}  &   19.7 (2.1e{-5})     &         285    (259, 317)                \\
                                       		& \multicolumn{1}{c|}{final} &           19.6 (2.2e{-5})            &     286    (260, 318)          \\ \midrule
\multirow{2}{*}{\makecell{N=15}}         	& \multicolumn{1}{c|}{abs.} &  62.2 (3.0e{-9})      & 1650    (1619, 1678)                          \\
                                       		& \multicolumn{1}{c|}{final} & 61.9 (6.5e{-9})      &1653    (1619, 1684)               \\\midrule
\multirow{2}{*}{\makecell{N=30}}  		& \multicolumn{1}{c|}{abs.} &  74.1 (3.7e{-9})    	&  6538    (6380, 6725)                        \\
                                       		& \multicolumn{1}{c|}{final} &  76.0 (2.6e{-9})       & 6518    (6362, 6695)            \\\midrule
\multirow{2}{*}{\makecell{N=100}} 		& \multicolumn{1}{c|}{abs.} & 5.82 (1.9e{-2})        & 36327    (27272, 50292)                           \\
                                        		& \multicolumn{1}{c|}{final} & 5.65 (1.8e{-2})        & 36618   (26339, 51579)             \\
\specialrule{.15em}{.05em}{.05em} 
\end{tabular}
\label{tb:t_test_heteroa}    \end{minipage} }
\end{tabular}
\vspace{-0.3cm}
\end{table}

\begin{figure*}[t]
\centering
\begin{tabular}{ccc}
\includegraphics[width=0.33\textwidth]{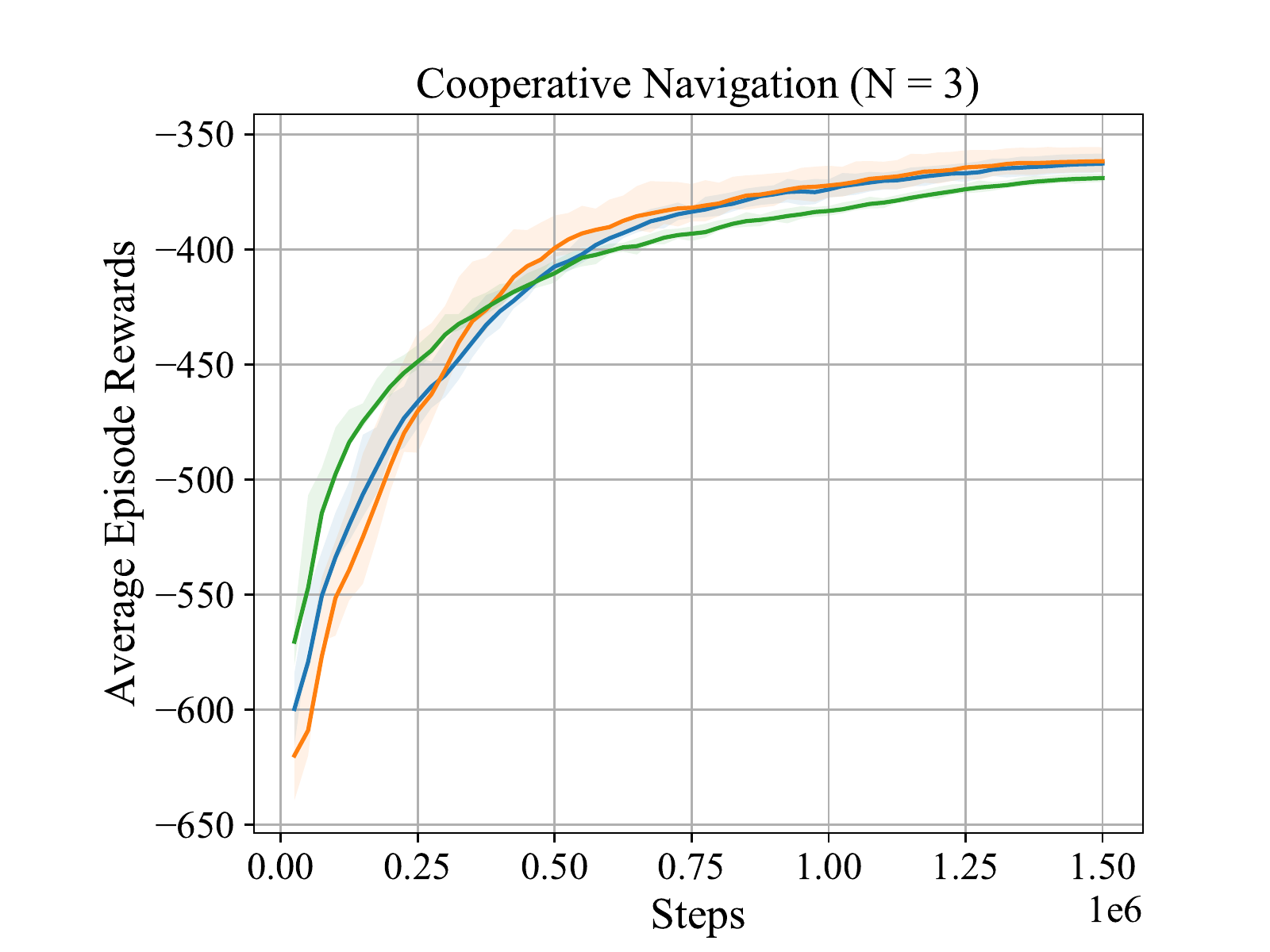}
&
\hspace{-0.6cm}\includegraphics[width=0.33\textwidth]{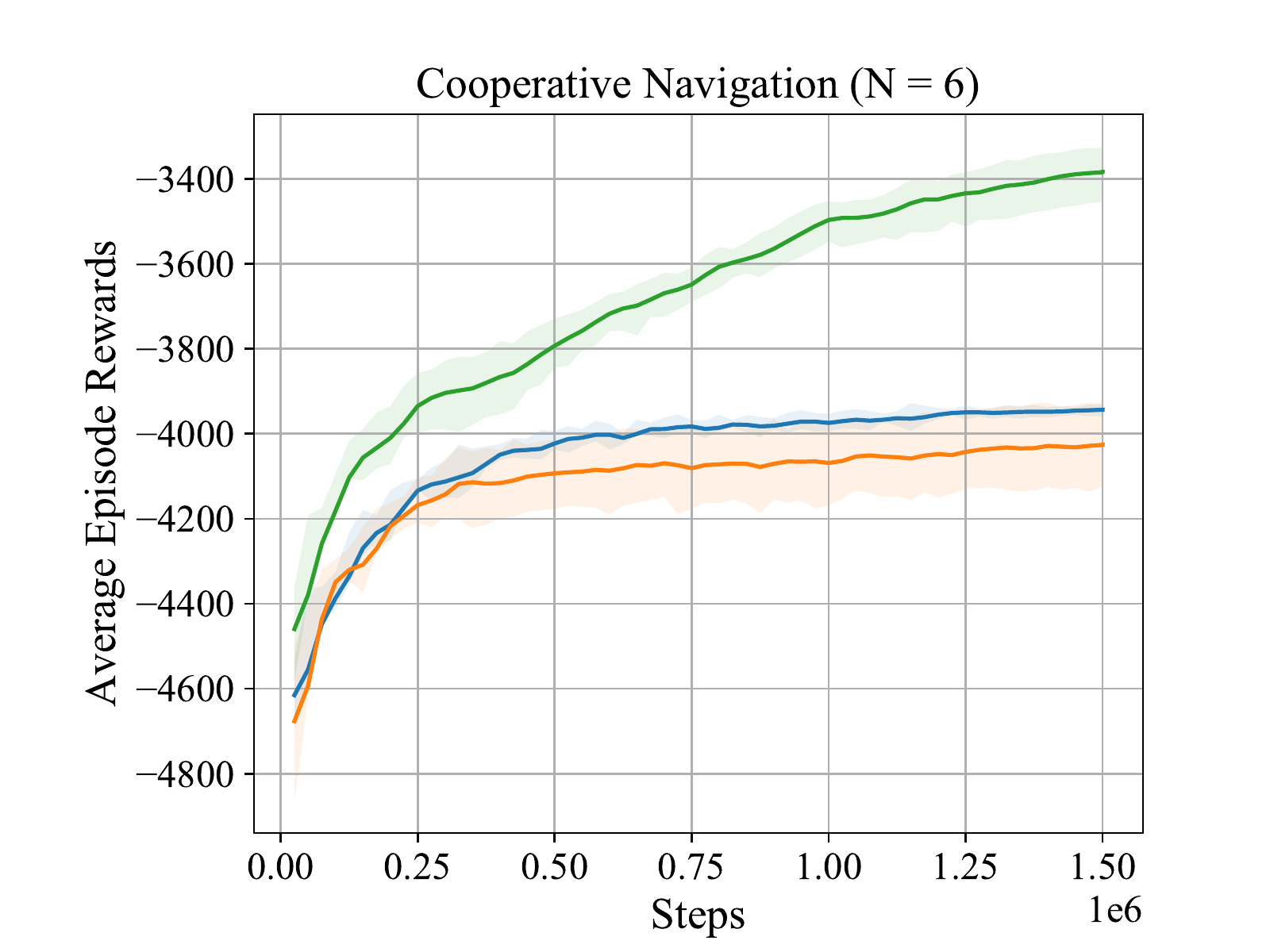}
&
\hspace{-0.6cm}\includegraphics[width=0.33\textwidth]{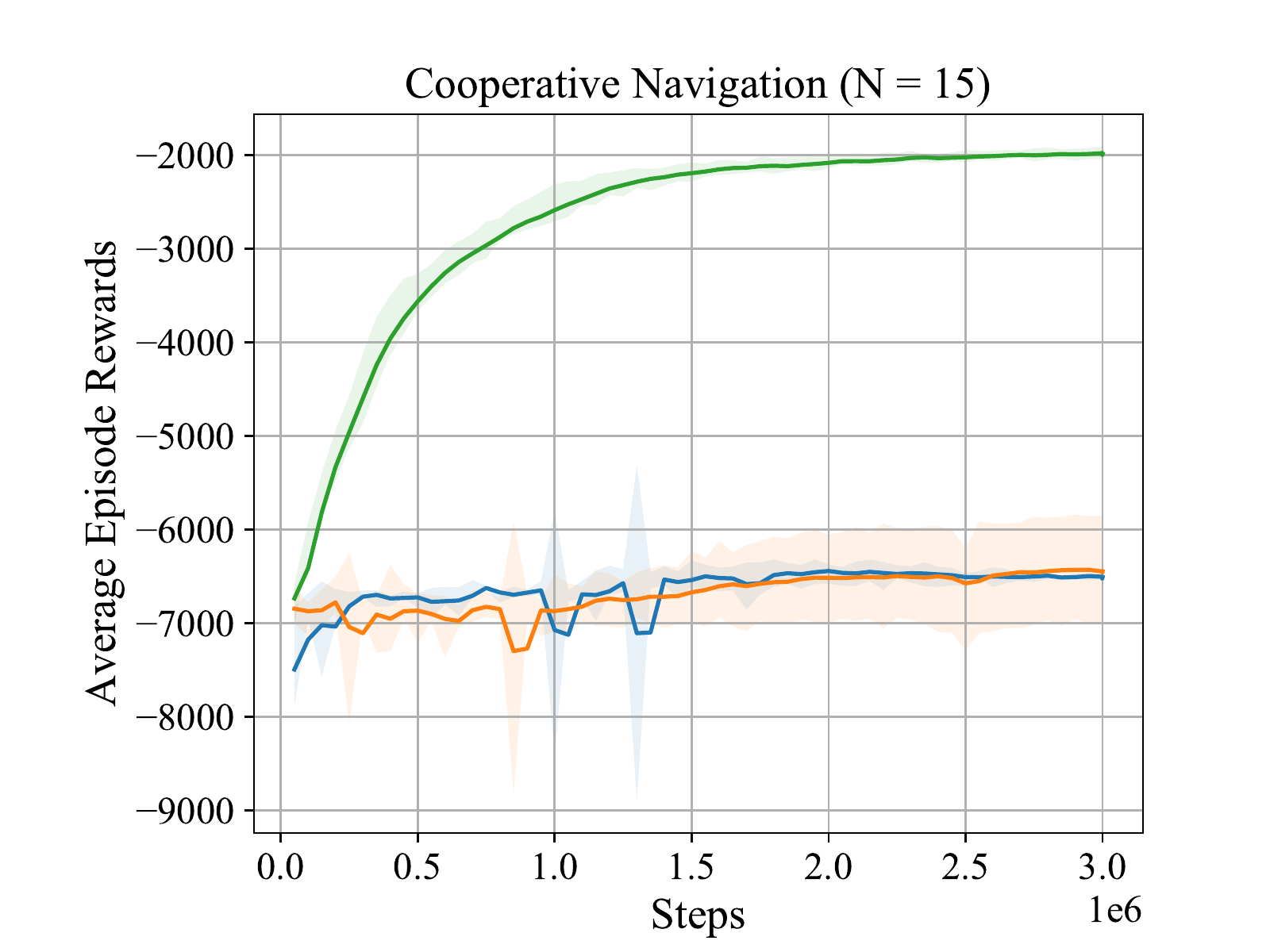}\\

\multicolumn{3}{c}{
\includegraphics[width=0.9\textwidth]{fig/legend.pdf}}\\
\end{tabular}
\vspace{-0.3cm}
\caption{Average episode reward comparison. Our permutation invariant critic (PIC) outperforms the MLP critic in all environment settings. }
\label{fig:r_plot_all1}
\vspace{-0.3cm}
\end{figure*}

\begin{figure*}[t]
\centering
\begin{tabular}{ccc}

\includegraphics[width=0.33\textwidth]{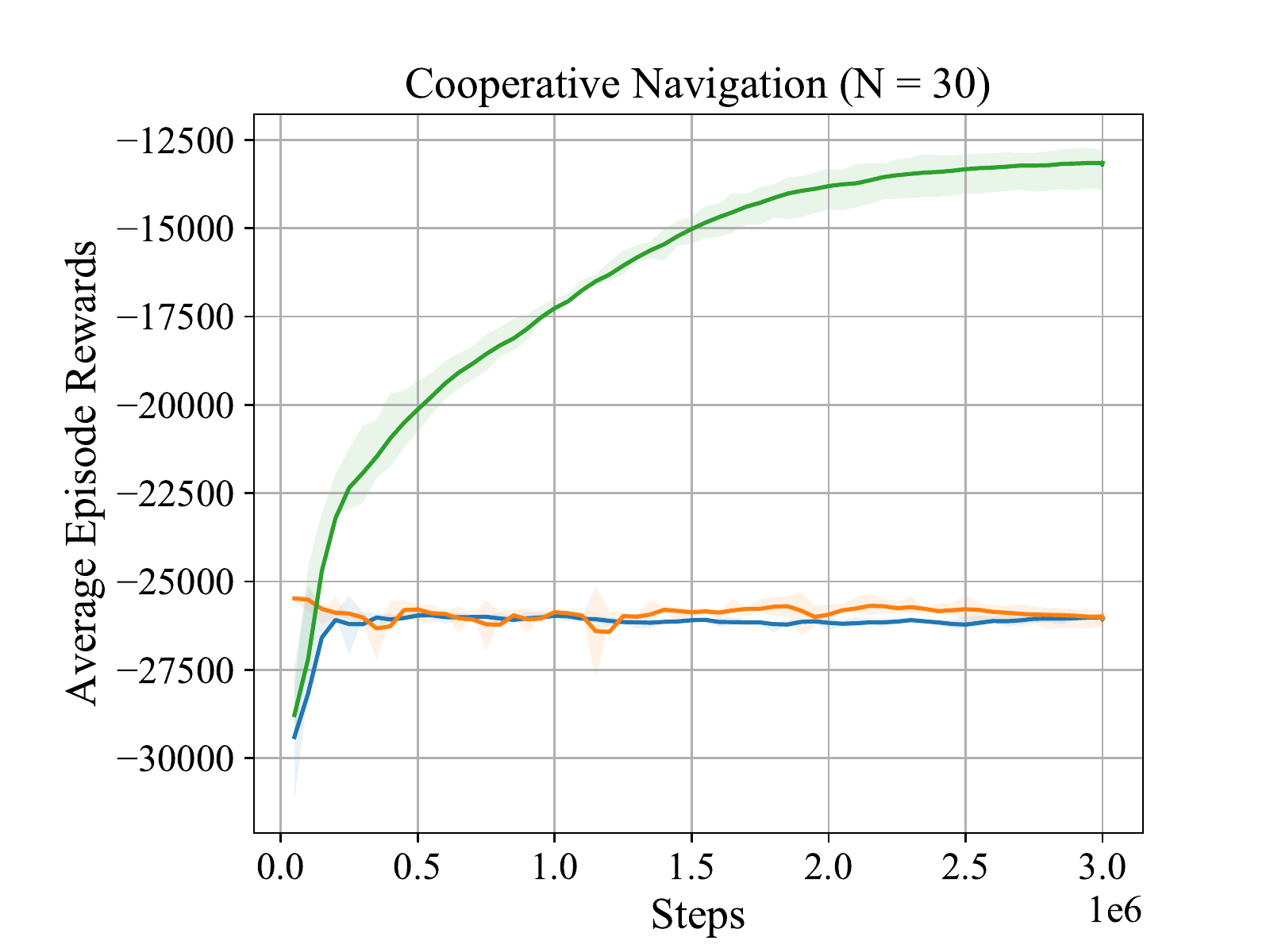}
&
\hspace{-0.6cm}\includegraphics[width=0.33\textwidth]{fig/Cooperative_Navigation_N_100_rewards.pdf}
&
\hspace{-0.6cm}\includegraphics[width=0.33\textwidth]{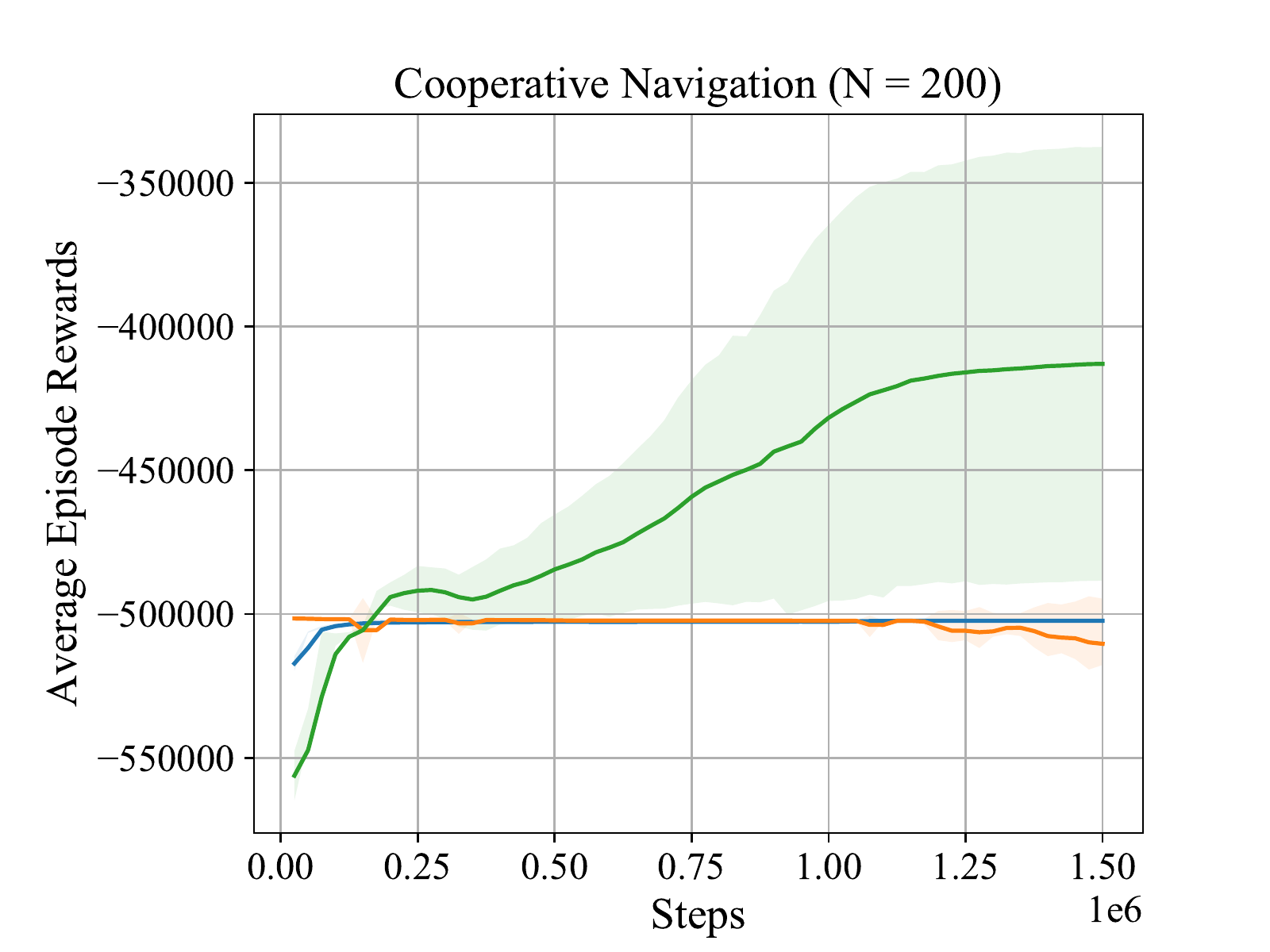}\\

\includegraphics[width=0.33\textwidth]{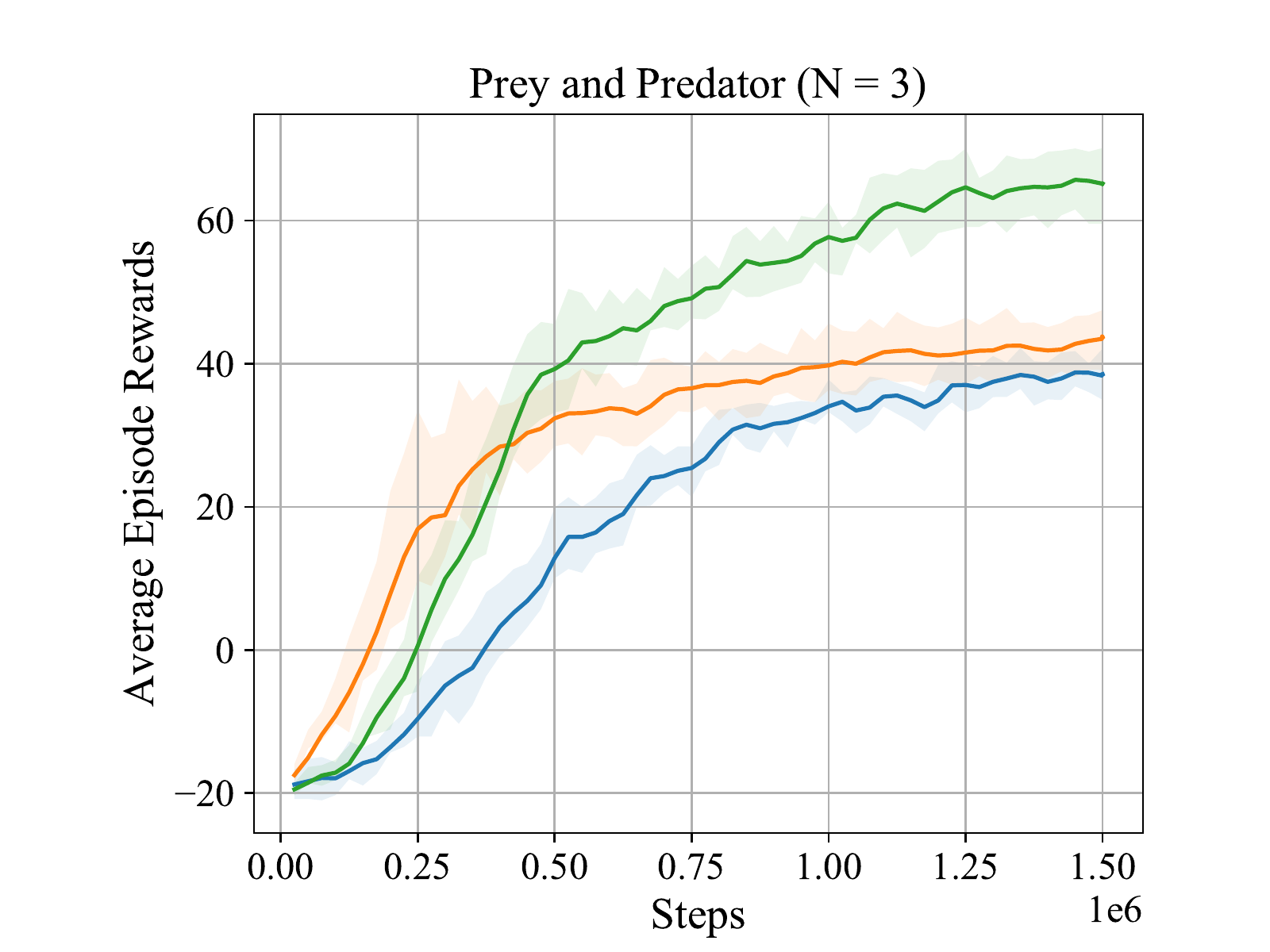}
&
\hspace{-0.6cm}\includegraphics[width=0.33\textwidth]{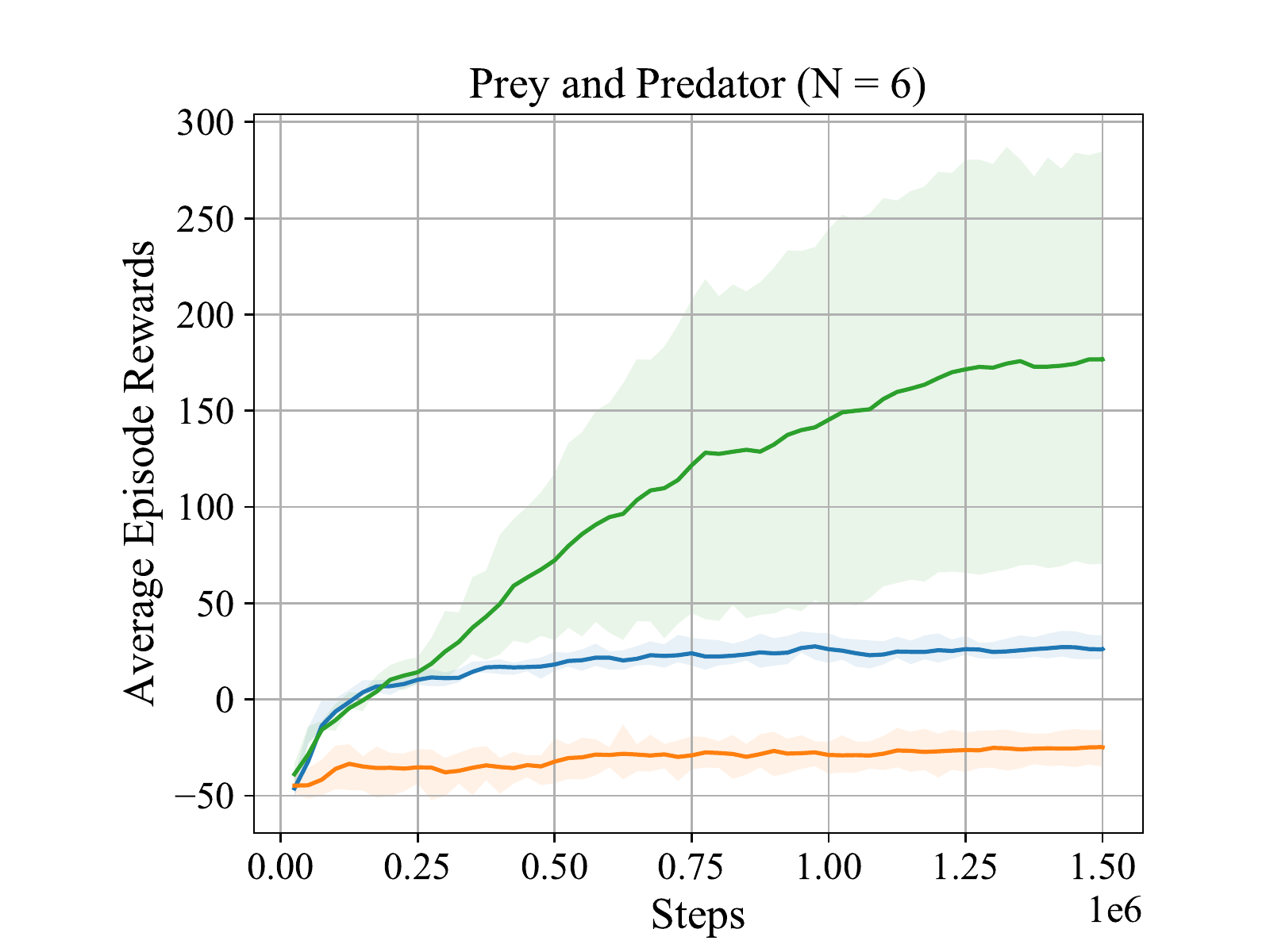}
&
\hspace{-0.6cm}\includegraphics[width=0.33\textwidth]{fig/Prey_and_Predator_N_15_rewards.pdf}\\

\includegraphics[width=0.33\textwidth]{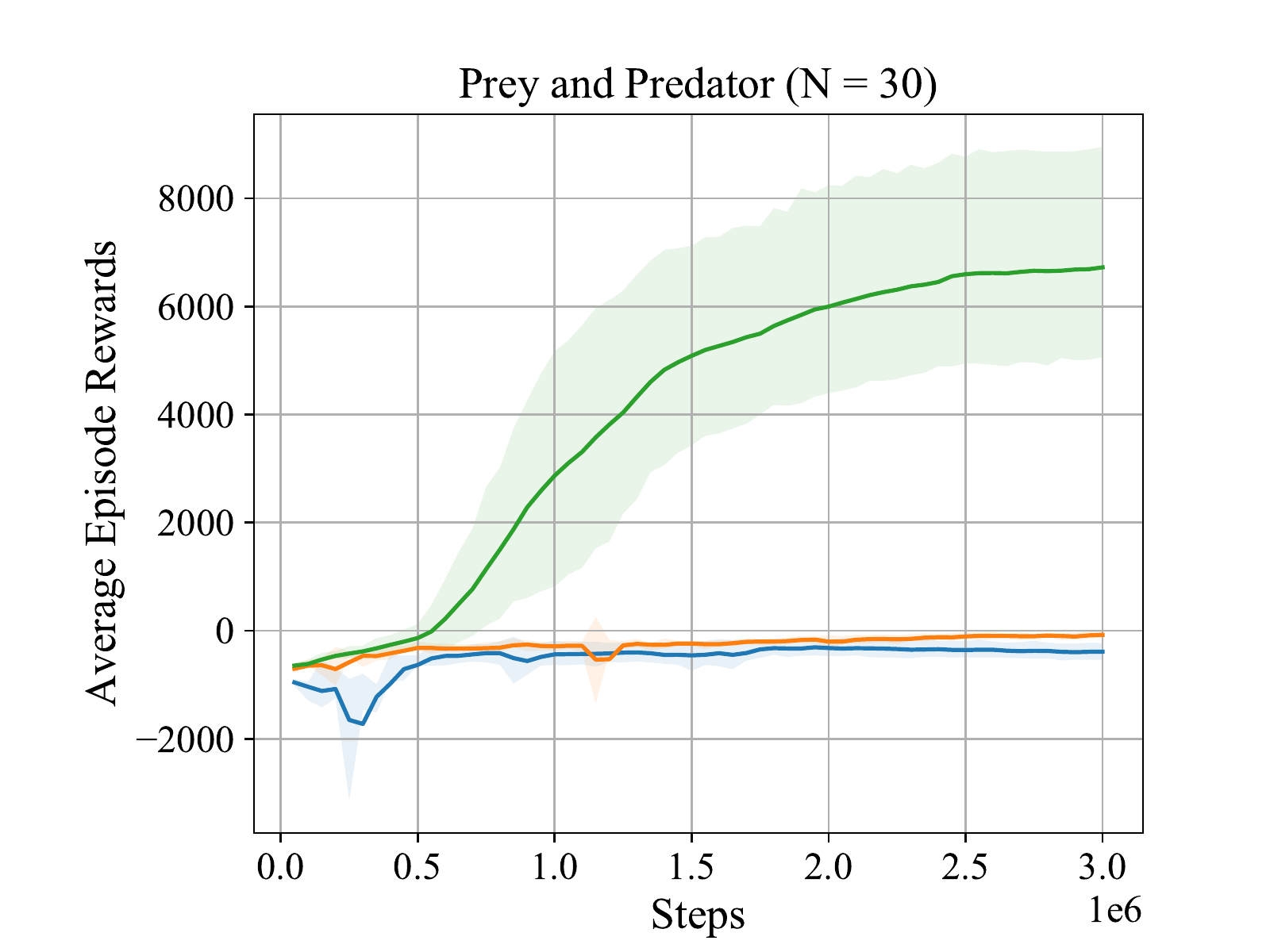}
&
\hspace{-0.6cm}\includegraphics[width=0.33\textwidth]{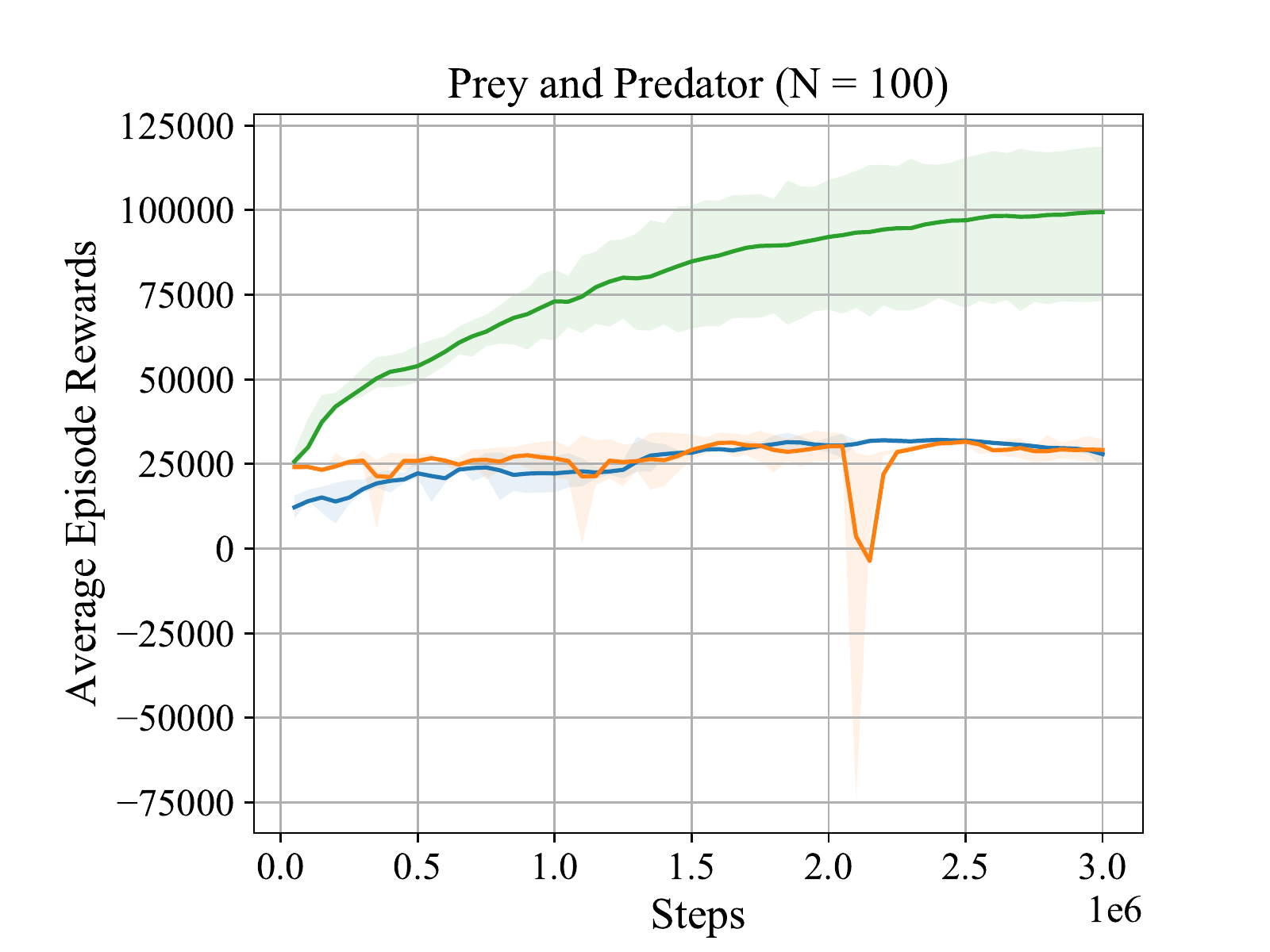}
&
\hspace{-0.6cm}\includegraphics[width=0.33\textwidth]{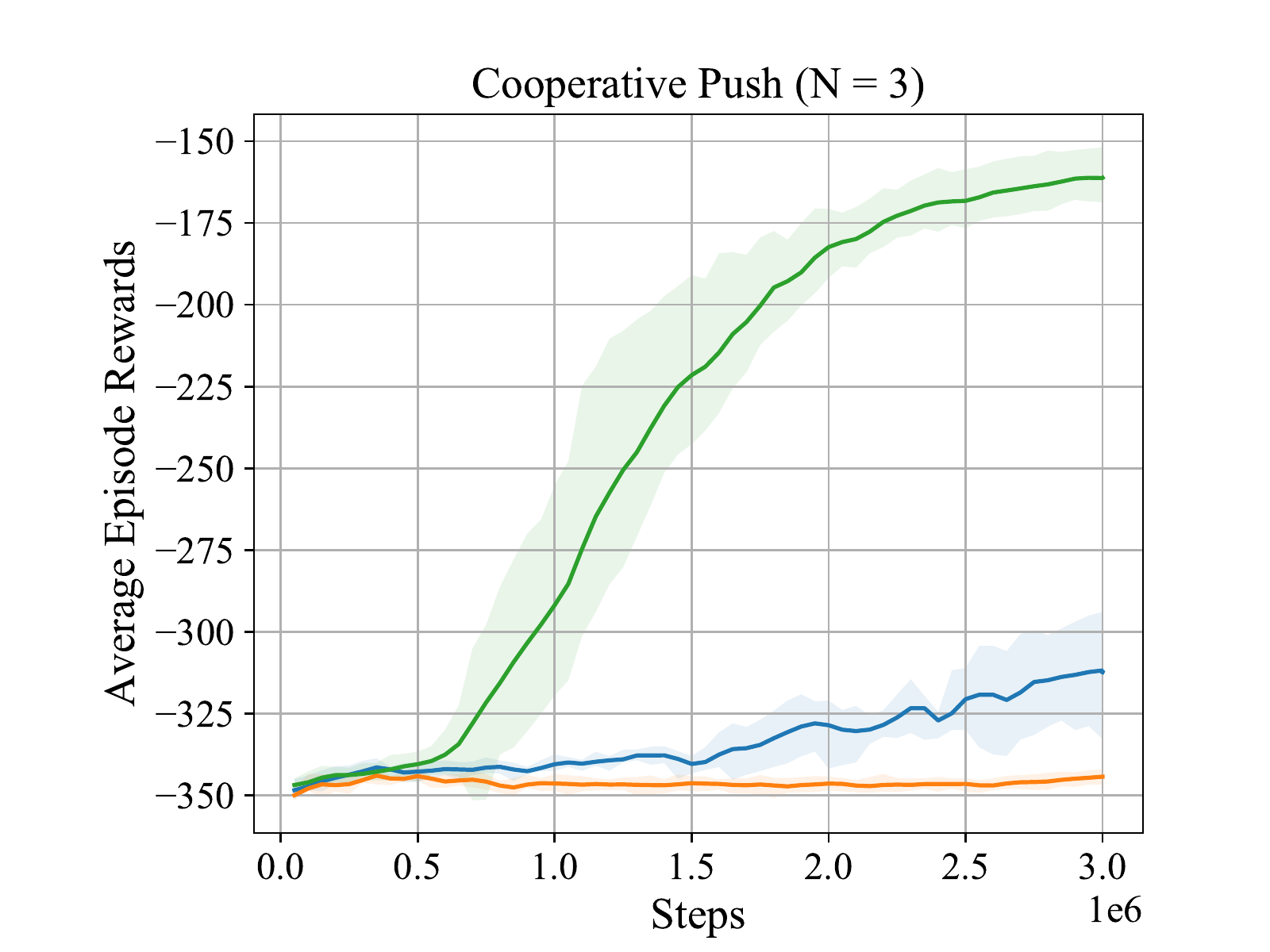}\\

\includegraphics[width=0.33\textwidth]{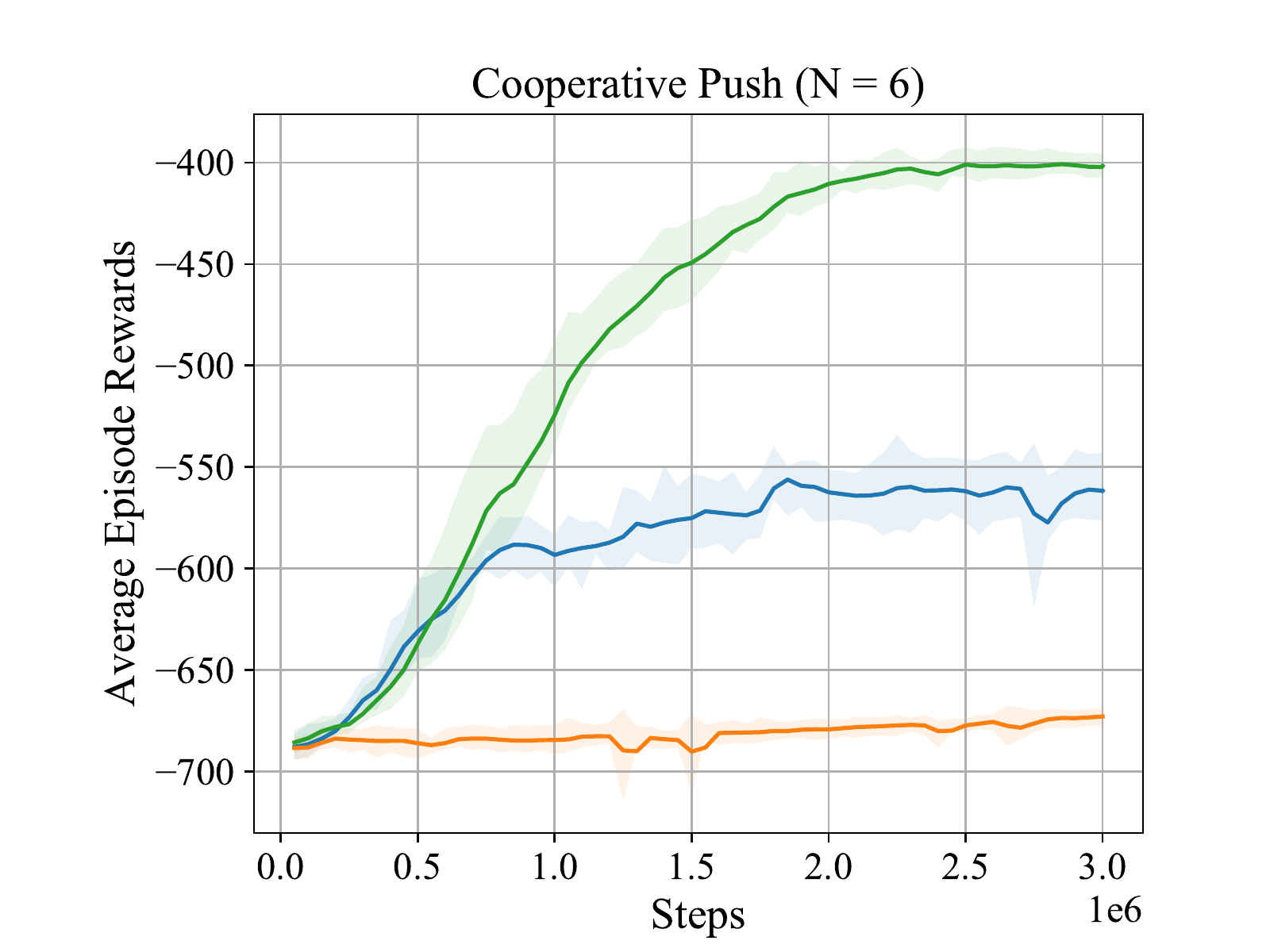}
&
\hspace{-0.6cm}\includegraphics[width=0.33\textwidth]{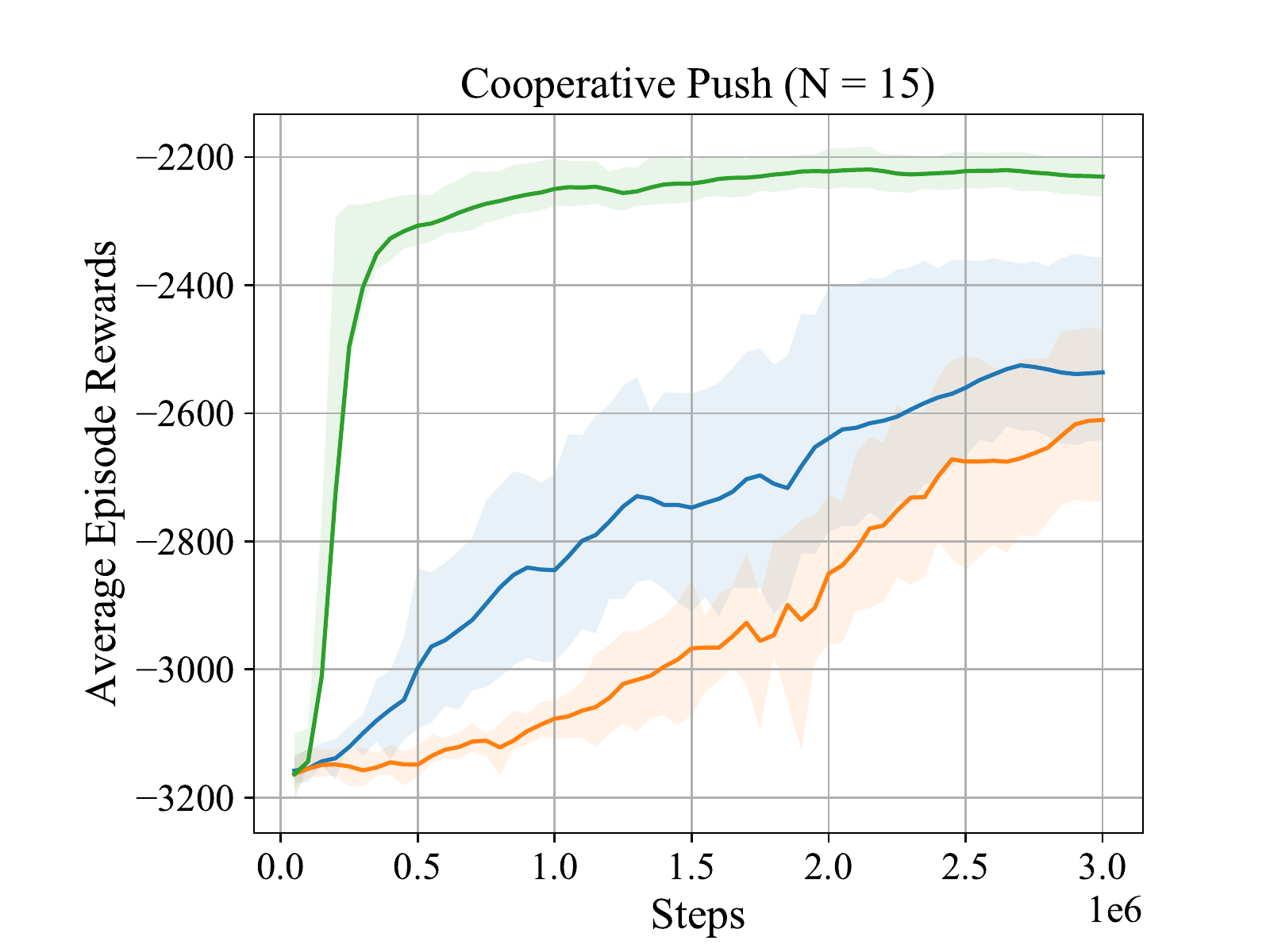}
&
\hspace{-0.6cm}\includegraphics[width=0.33\textwidth]{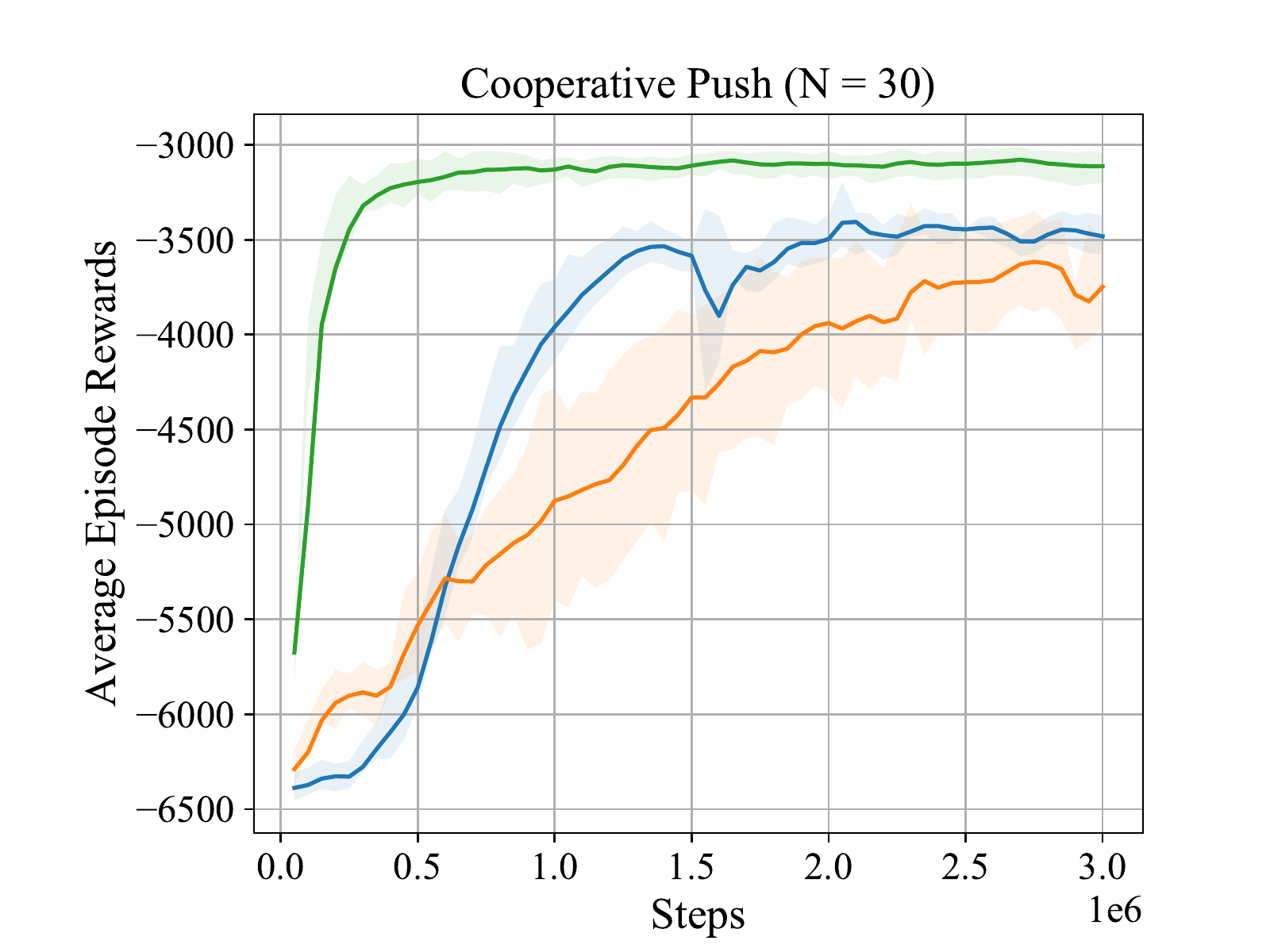}\\

\includegraphics[width=0.33\textwidth]{fig/Heterogeneous_Navigation_N_4_rewards.pdf}
&
\hspace{-0.6cm}\includegraphics[width=0.33\textwidth]{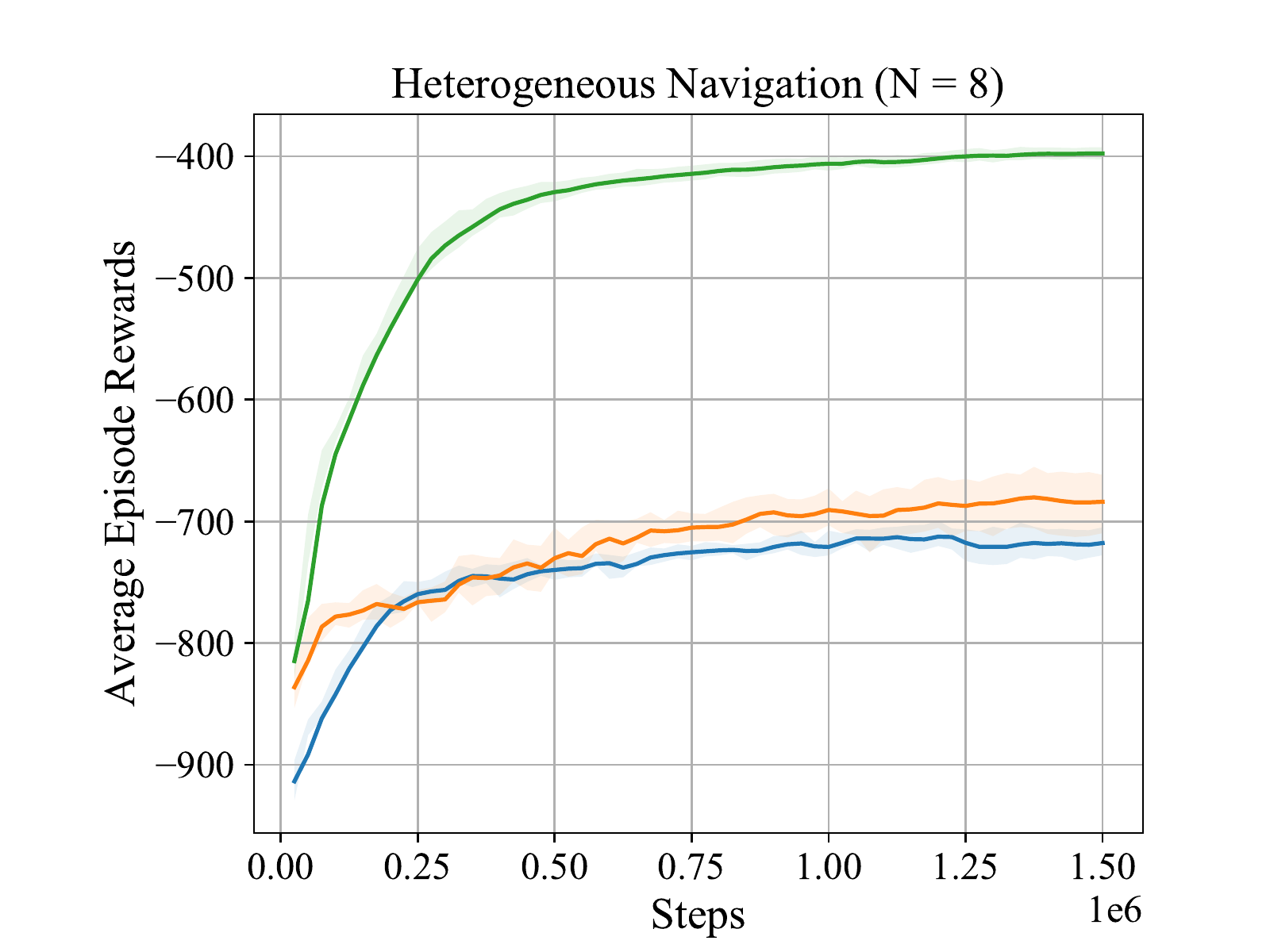}
&
\hspace{-0.6cm}\includegraphics[width=0.33\textwidth]{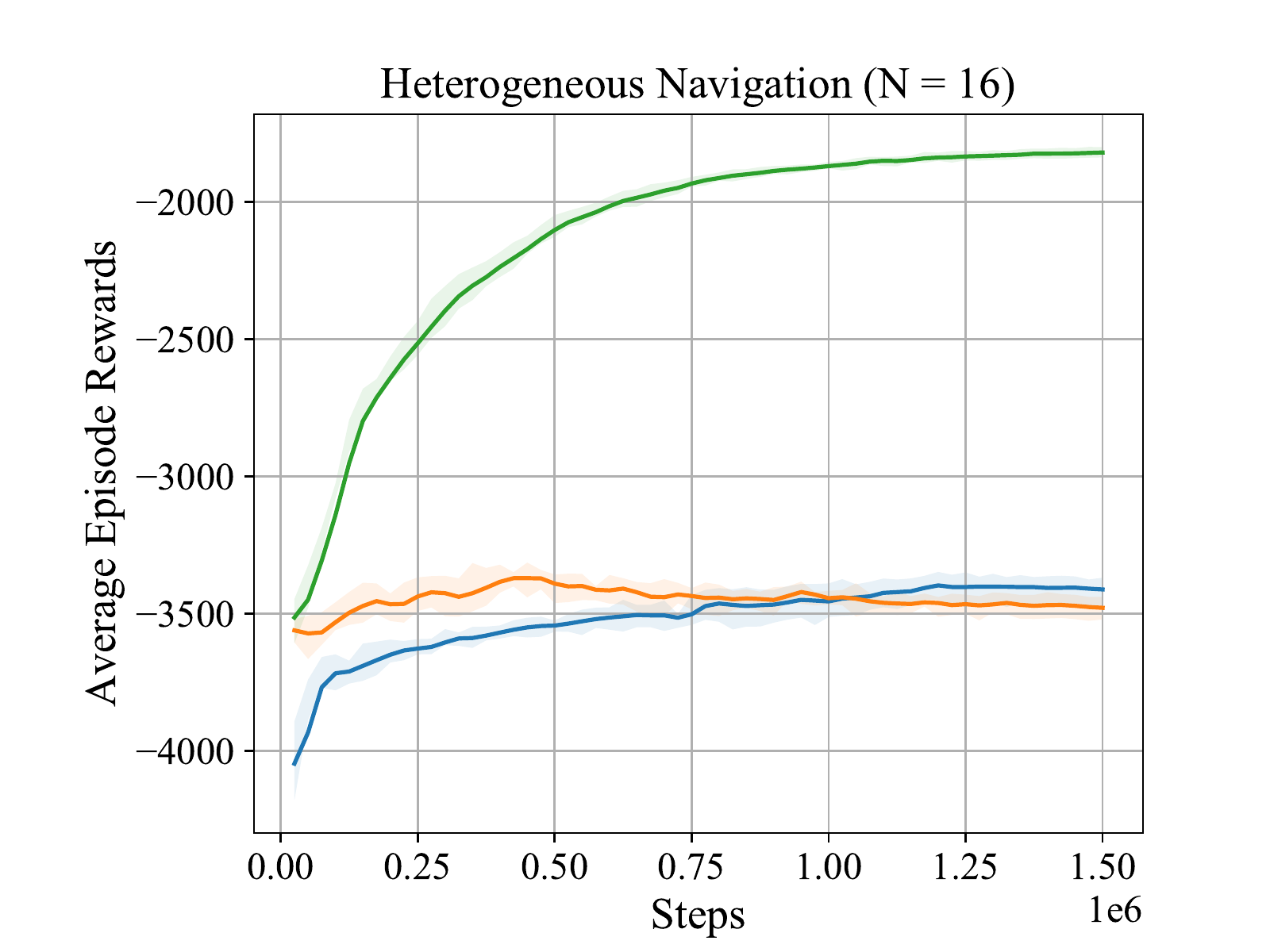}\\
\includegraphics[width=0.33\textwidth]{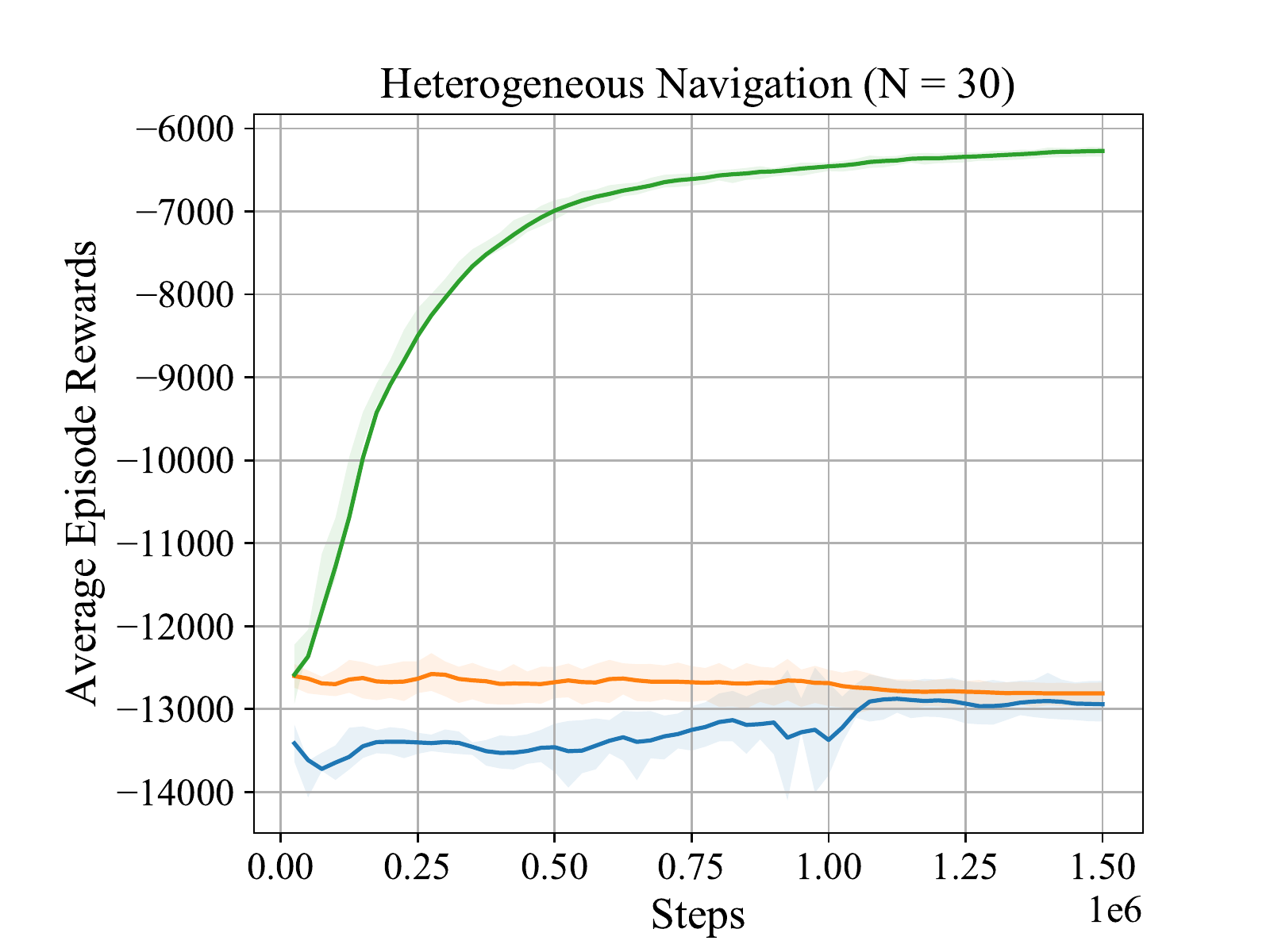}
&
\hspace{-0.6cm}\includegraphics[width=0.33\textwidth]{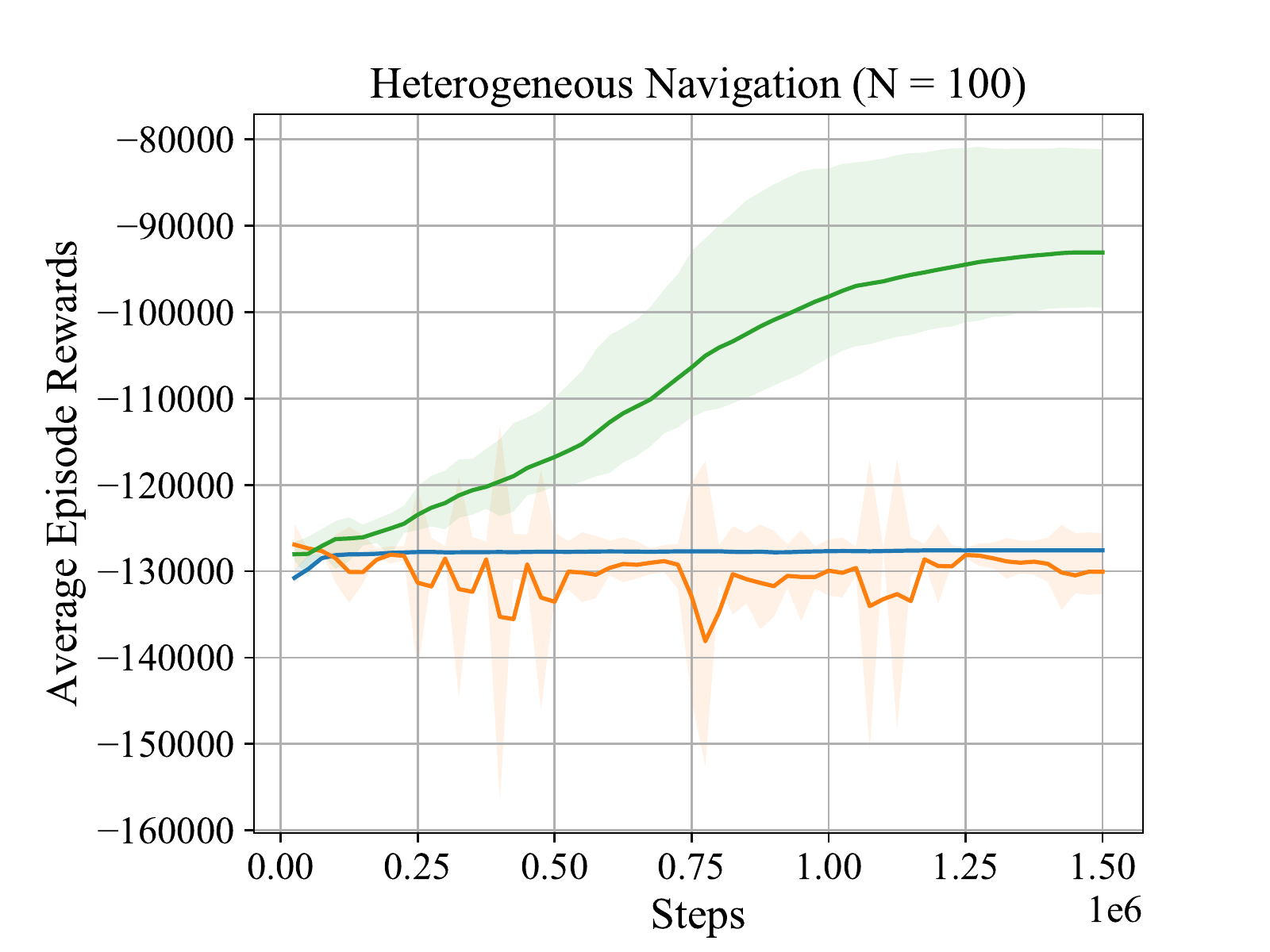}\\
\multicolumn{3}{c}{
\includegraphics[width=0.9\textwidth]{fig/legend.pdf}}\\
\end{tabular}
\vspace{-0.3cm}
\caption{Average episode reward comparison. Our permutation invariant critic (PIC) outperforms the MLP critic in all environment settings. 
}
\label{fig:r_plot_all2}
\vspace{-0.3cm}
\end{figure*}